\documentclass[twoside,11pt]{article}

\usepackage{jmlr2e}

\usepackage[table, svgnames, dvipsnames]{xcolor}
\usepackage[utf8]{inputenc} 
\usepackage{hyperref}       
\usepackage{natbib}
\usepackage{amsmath}
\usepackage{mathtools}
\usepackage{amssymb}
\usepackage{subfigure}
\usepackage{booktabs}
\usepackage{caption}
\usepackage[capitalise,nameinlink]{cleveref}
\usepackage[normalem]{ulem}

\newcommand{\updated}[1]{#1}

\newcommand{\natgrad}{\widetilde{\nabla}}
\newcommand{\grad}{\nabla}

\newcommand{\entropy}{\mathcal{H}}
\newcommand{\barloss}{\bar{\ell}}

\newcommand{\loss}{\ell}

\newcommand{\vparam}{\vtheta}
\newcommand{\param}{\theta}

\newcommand{\tvparam}{\tilde{\vparam}}

\newcommand{\minibatch}{\mathcal{M}}

\newcommand{\natparam}{\lambda}
\newcommand{\vnatparam}{\vlambda}

\newcommand{\meanparam}{\mu}
\newcommand{\vmeanparam}{\vmu}

\newcommand{\lat}{z}
\newcommand{\vlat}{\vz}

\newcommand{\dkls}[3]{\mathbb{D}_{KL}^{#1}[#2 \, \|\, #3]}

\newcommand{\deriv}[2]{\frac{\partial{#1}}{\partial{#2}}}

\newcommand\cut[1]{}

\newcommand{\elbofinal}{\mathcal{L}}

\newcommand{\tvm}{\widetilde{\vm}}

\newcommand{\tvlambda}{\widetilde{\vlambda}}

\newcommand{\squishlist}{
   \begin{list}{$\bullet$}
    { \setlength{\itemsep}{0pt}      \setlength{\parsep}{3pt}
      \setlength{\topsep}{3pt}       \setlength{\partopsep}{0pt}
      \setlength{\leftmargin}{1.5em} \setlength{\labelwidth}{1em}
      \setlength{\labelsep}{0.5em} } }

\newcommand{\squishlisttwo}{
   \begin{list}{$\bullet$}
    { \setlength{\itemsep}{0pt}    \setlength{\parsep}{0pt}
      \setlength{\topsep}{0pt}     \setlength{\partopsep}{0pt}
      \setlength{\leftmargin}{2em} \setlength{\labelwidth}{1.5em}
      \setlength{\labelsep}{0.5em} } }

\newcommand{\squishend}{
    \end{list}  }

{}
{}
{}
{}

\newcommand{\half}{\text{$\frac{1}{2}$}}

\newcommand{\real}{\text{$\mathbb{R}$}}

\newcommand{\rnd}[1]{\left(#1\right)}
\newcommand{\sqr}[1]{\left[#1\right]}
\newcommand{\crl}[1]{\left\{#1\right\}}
\newcommand{\myang}[1]{\langle#1\rangle}
\newcommand{\myexpect}{\mathbb{E}}

\newcommand{\gauss}{\text{${\cal N}$}}

\newcommand{\myvec}[1]{\ensuremath{\boldsymbol{#1}}}
\newcommand{\myvecsym}[1]{\ensuremath{\boldsymbol{#1}}}

\newcommand{\vzero}{\text{$\myvecsym{0}$}}
\newcommand{\vone}{\text{$\myvecsym{1}$}}

\newcommand{\valpha}{\text{$\myvecsym{\alpha}$}}

\newcommand{\vdelta}{\text{$\myvecsym{\delta}$}}

\newcommand{\vepsilon}{\text{$\myvecsym{\epsilon}$}}

\newcommand{\vmu}{\text{$\myvecsym{\mu}$}}

\newcommand{\vlambda}{\text{$\myvecsym{\lambda}$}}

\newcommand{\vtheta}{\text{$\myvecsym{\theta}$}}

\newcommand{\va}{\text{$\myvec{a}$}}
\newcommand{\vb}{\text{$\myvec{b}$}}

\newcommand{\vf}{\text{$\myvec{f}$}}

\newcommand{\vm}{\text{$\myvec{m}$}}

\newcommand{\vs}{\text{$\myvec{s}$}}

\newcommand{\vu}{\text{$\myvec{u}$}}
\newcommand{\vv}{\text{$\myvec{v}$}}

\newcommand{\vx}{\text{$\myvec{x}$}}

\newcommand{\vy}{\text{$\myvec{y}$}}

\newcommand{\vz}{\text{$\myvec{z}$}}
\newcommand{\vA}{\text{$\myvec{A}$}}

\newcommand{\vC}{\text{$\myvec{C}$}}

\newcommand{\vF}{\text{$\myvec{F}$}}

\newcommand{\vH}{\text{$\myvec{H}$}}
\newcommand{\vI}{\text{$\myvec{I}$}}

\newcommand{\vS}{\text{$\myvec{S}$}}
\newcommand{\vT}{\text{$\myvec{T}$}}

\newcommand{\vX}{\text{$\myvec{X}$}}

\newcommand{\diag}{\text{{diag}}}

\newcommand{\trace}{\text{Tr}}

\newcommand{\calD}{\text{${\cal D}$}}

\newcommand{\data}{\calD}

\newcommand{\be}{\begin{equation}}
\newcommand{\ee}{\end{equation}}
\newcommand{\bea}{\begin{eqnarray}}
\newcommand{\eea}{\end{eqnarray}}
\newcommand{\beaa}{\begin{eqnarray*}}
\newcommand{\eeaa}{\end{eqnarray*}}

\DeclareMathOperator*{\argmin}{arg\,min}
\crefname{section}{Sec.}{Sections}
\crefname{appendix}{App.}{Appendices}
\crefname{algorithm}{Alg.}{Algorithms}
\crefname{equation}{Eq.}{Eqs.}
\crefname{figure}{Fig.}{Figures}
\creflabelformat{equation}{#2\textup{#1}#3} 

\usepackage{lastpage}
\jmlrheading{24}{2023}{1-\pageref{LastPage}}{3/22; Revised
6/23}{9/23}{22-0291}{Mohammad Emtiyaz Khan and H{\aa}vard Rue}
\ShortHeadings{The Bayesian Learning Rule}{Khan and Rue}
\firstpageno{1}

\begin{document}

\title{The Bayesian Learning Rule}

\author{\name Mohammad Emtiyaz Khan \email emtiyaz.khan@riken.jp \\
       \addr RIKEN Center for Advanced Intelligence Project\\
       1-4-1 Nihonbashi, Chuo-ku, Tokyo 103-0027, Japan 
       \AND
       \name H{\aa}vard Rue \email haavard.rue@kaust.edu.sa \\
       \addr CEMSE Division\\
       King Abdullah University of Science and Technology\\
       Thuwal 23955-6900, Saudi Arabia}

\editor{Justin Domke}

\maketitle

\begin{abstract}%
    We show that many machine-learning algorithms are specific
    instances of a single algorithm called the \emph{Bayesian learning
        rule}. The rule, derived from Bayesian principles, yields a
    wide-range of algorithms from fields such as optimization, deep
    learning, and graphical models. This includes classical algorithms
    such as ridge regression, Newton's method, and Kalman filter, as
    well as modern deep-learning algorithms such as
    stochastic-gradient descent, RMSprop, and Dropout. The key idea in
    deriving such algorithms is to approximate the posterior using
    candidate distributions estimated by using natural gradients.
    Different candidate distributions result in different algorithms
    and further approximations to natural gradients give rise to
    variants of those algorithms. Our work not only unifies,
    generalizes, and improves existing algorithms, but also helps us
    design new ones.
\end{abstract}


\section{Introduction}
\label{sec:intro}
\subsection{Learning-Algorithms}

Machine Learning (ML) methods have been extremely successful in
solving many challenging problems in fields such as computer vision,
natural-language processing, and artificial intelligence. The main
idea is to formulate those problems as prediction problems and learn
a model on existing data to predict the future outcomes. For example,
to recognize objects in images, we 
collect $N$ images $\vx_i\in\real^D$ and object labels
$y_i\in\{1,2,\ldots,K\}$, and learn a model $f_{\text{\vparam}}(\vx)$
with parameters $\vparam \in\real^P$ to predict the label for a new
image. Learning algorithms are often employed to estimate 
$\vparam$ by Empirical Risk Minimization (ERM),
\begin{equation}
    \vparam_* = \arg\min_{\text{\vparam}} \barloss(\vparam) 
    \quad\text{where}\quad
    \barloss(\vparam) = 
    \sum_{i=1}^N \loss( y_i, f_{\text{\vparam}}(\vx_i) ) + R(\vparam).
    \label{eq:ERM}
\end{equation}
Here, $\loss(y,f_{\text{\vparam}}(\vx))$ is a loss function that
encourages the model to predict well and ${R}(\vparam)$ is a
regularizer that prevents it from overfitting. A wide-variety of such
learning-algorithms exist in the literature to solve a variety of
learning problems, for example, ridge regression, Kalman filters,
gradient descent, and Newton’s method. These algorithms play a key
role in the success of modern ML.

Learning-algorithms are often derived by borrowing and combining ideas
from a diverse set of fields, such as statistics, optimization, and
computer science. For example, the field of probabilistic graphical
models~\citep{koller2009probabilistic, bishop2006pattern} uses popular
algorithms such as ridge-regression~\citep{hoerl1970ridge}, Kalman
filters~\citep{Kalman1960ANA}, the Viterbi algorithm for
Hidden Markov Models~\citep{stratonovich1965conditional}, and
Expectation-Maximization (EM)~\citep{dempster1977maximum}. The field
of approximate inference builds upon such algorithms to
perform inference on complex graphical models, for example, algorithms such
as Laplace's
method~\citep{laplace1986memoir,art367,art375,art451,art632},
stochastic variational inference (SVI)~\citep{hoffman2013stochastic,
    sato2001online}, Variational message passing
(VMP)~\citep{winn2005variational} etc. Similarly, the field of
continuous optimization has its own popular methods such as gradient
descent~\citep{cauchy1847methode}, Newton's method, and mirror
descent~\citep{nemirovski1978cesaro}, and deep-learning
(DL) methods use them to design new techniques 
that work with massive models and data sets, for example,
stochastic-gradient descent~\citep{robbins1951},
RMSprop~\citep{hintonTieleman}, Adam~\citep{kingma2014adam}, Dropout
regularization~\citep{Srivastava2014}, and Straight-Through Estimator
(STE)~\citep{bengio2013estimating}. Such mixing of algorithms from
diverse fields is a strength of the ML community, and our goal here is
to provide common principles to unify, generalize, and improve such algorithms.

\subsection{The Bayesian Learning Rule}

We show that a wide-range of well-known learning-algorithms from a
variety of fields are all specific instances of a single learning
algorithm derived from Bayesian principles. \updated{The starting point is to extend 
\cref{eq:ERM} by using the variational formulation of \cite{art669}, where we optimize over a well-defined candidate
distribution $q(\vparam)$, and for which the minimizer
\begin{equation}
  q_*(\vparam) = \argmin_{q(\text{\vparam})}  \,\,
   \myexpect_{q} \sqr{  \sum_{i=1}^N
    \loss(y_i, f_{\text{\vparam}}(\vx_i)) }
    + \dkls{}{q(\vparam)}{p(\vparam)}
    \label{eq:BayesP}
\end{equation}}
defines a generalized posterior \citep{10.1145/307400.307433, catoni2007pac, art670} in lack of a precise
likelihood. 
\updated{We can rewrite the objective in terms of entropy $\mathcal{H}(q)=\myexpect_{q}[-\log q(\vparam)]$ by expanding the Kullback-Leibler Divergence (KLD) as follows,
\[
   \dkls{}{q(\vparam)}{p(\vparam)} = \myexpect_q \sqr{ \log \frac{q(\vparam)}{p(\vparam)} } = \myexpect_q \sqr{ R(\vparam) } - \mathcal{H}(q) ,\]
where we chose the prior to be $p(\vparam)\propto \exp(-R(\vparam))$.
Plugging this in \cref{eq:BayesP}, we get 
\[ q_*(\vparam) = \argmin_{q(\text{\vparam})} \,\, \myexpect_{q} \rnd{  \barloss(\vparam) } - \entropy(q). \] }
When 
$\exp(- \loss(y_i, f_{\text{\vparam}}(\vx_i)))$ is proportional to the
likelihood of $y_i$ for all $i$, then $q_*(\vparam)$ is the
posterior distribution \citep{art669}; see
\cref{app:bayes_as_opt} \updated{for a proof}.

\updated{We use two components to derive learning algorithms from this Bayesian
formulation,} 
\begin{enumerate}
\item A (sub-)class of distributions ${\mathcal Q}$ to optimize over.
    In our discussion, we will assume ${\mathcal Q}$ to be the set of a
      regular and minimal exponential family
    \begin{displaymath}
        q_\vnatparam(\vparam) = h(\vparam) \exp\sqr{\myang{ \vnatparam,
                \vT(\vparam) } - A(\vnatparam)},
        \label{eq:exp_fam}
    \end{displaymath}
    where $\vnatparam \in \Omega \subset \real^M$ are the natural (or
    canonical) parameter in a non-empty open set $\Omega$. 
    The cumulant (or log partition) function $A(\vnatparam)$ is
    finite, strictly convex and differentiable over $\Omega$. Further,
    $\vT(\vparam)$ is the sufficient statistics, $\myang{\cdot,\cdot}$
    is an inner product and $h(\vparam)$ is some function. The
    expectation parameter 
      $\vmeanparam(\vnatparam) = \myexpect_{q_\vnatparam} [\vT(\vparam)] \in \mathcal{M}$ is 
    a (bijective) function of $\vnatparam$. Examples later will
    include the multivariate Normal distribution and the Bernoulli
      distribution \updated{(see \cref{tab:EFsummary} for a list).}
    
\item An optimizing algorithm, called the \emph{Bayesian learning
   rule} (BLR), that locates the best candidate $q_{\vnatparam_*}(\vparam)$ in
      ${\mathcal Q}$, by updating the candidate $q_{\vnatparam_t}(\vparam)$ with natural
      parameter $\vnatparam_t$ at iteration $t$ \updated{using a sequence of learning rates $\rho_t>0$}:
    \begin{equation}
        \vnatparam_{t+1} \leftarrow \vnatparam_t -
        \rho_t \natgrad_\vnatparam \sqr{ 
        \myexpect_{q_{\vnatparam_t}}
        \rnd{\barloss(\vparam)} - \entropy(q_{\vnatparam_t}) }.
        \label{eq:bayes_learn_rule_general}
    \end{equation}
      The updates use
    the \emph{natural-gradients}~\citep{amari1998natural}, denoted by
      $\natgrad_\vnatparam$, and defined as
      \begin{equation}
         \natgrad_\vnatparam \myexpect_{q_{\vnatparam_t}}(\cdot)   
         = \vF(\vnatparam_t)^{-1} \,
         \sqr{ \left. \grad_{\vnatparam} \myexpect_{q_\vnatparam}(\cdot)
             \right\vert_{\text{\vnatparam} = \text{\vnatparam}_t} }
         \,\,= \,\,\left. \grad_{\vmeanparam} \myexpect_{q_\vnatparam}(\cdot)
         \right\vert_{\vmeanparam = \grad_\vnatparam A(\vnatparam_t)},
         \label{eq:natgrad_def}
      \end{equation}
      which rescale the vanilla gradients
      $\grad_\vnatparam$ with the Fisher information matrix (FIM)
      $\vF(\vlambda) = \grad_{\vnatparam}^2 A(\vnatparam)$
      to adjust for the geometry in the natural-parameter space $\Omega$.
      The second equality
      follows from the chain rule to express natural-gradients as
      vanilla gradients with respect to
      $\vmu = \nabla_\vnatparam A(\vnatparam)$~\citep{malago2011towards, raskutti2015information}. Throughout, we will
      use this property to simplify natural-gradient computations.
      These details are discussed in \cref{sec:bayes_learn}, where we
      show that the BLR update can also be seen as a mirror descent
      algorithm, where the geometry of the Bregman divergence is
      dictated by the chosen exponential-family. Throughout, we will assume
      $\vnatparam_{t}\in \Omega$ for all $t$, which in practice might
      require a line-search or a projection step.
\end{enumerate}
The main message of this paper is that many well-known
learning-algorithms -- such as those used in optimization, deep
learning, and machine learning in general -- can be derived directly
following the above scheme via a single algorithm (BLR) that
optimizes \cref{eq:BayesP}. Different exponential families
${\mathcal Q}$ give rise to different algorithms, and within those,
various approximations to natural-gradients that are needed, give rise
to many variants. These results are extended to mixture-candidates
in \cref{sec:multimodal}, and we expect
    them to hold \updated{in general for other classes of candidate distributions}.

Our use of natural-gradients here is not a matter of choice. In fact,
natural-gradients are inherently present in \emph{all} solutions of the
    Bayesian objective in \cref{eq:BayesP}. \updated{For example, the natural parameter $\vnatparam_*$ of a solution $q_{\vnatparam_*}(\vparam)$ is a fixed point of \cref{eq:bayes_learn_rule_general}, and at convergence
\begin{equation}
   \begin{split}
      \vnatparam_*  =\vnatparam_* - \rho_* \natgrad_\vnatparam \sqr{ \myexpect_{q_{\vnatparam_*}} \sqr{\barloss(\vparam)} - \entropy(q_{\vnatparam_*}) }
      &\implies
       \vnatparam_* = \natgrad_{\vnatparam} \myexpect_{q_{\vnatparam_*}}\sqr{-\barloss(\vparam)}.
   \end{split}
   \label{eq:fixed_pt_specific}
\end{equation}
In the second equality, we assume that $h(\vparam)$ is a constant, so that $\natgrad_{\vnatparam} \mathcal{H}(q_\vnatparam) = -\vnatparam$
(see \cref{app:natgrad_entropy}). The equality states that the natural parameter $\vnatparam_*$ must be equal to the natural gradient of $\myexpect_{q_{\vnatparam_*}}\sqr{-\barloss(\vparam)}$, therefore estimation of $q_{\vnatparam_*}(\vparam)$ is tightly coupled to computation of natural gradients. Depite this,} the importance of natural-gradients is entirely missed
in the Bayesian/variational inference literature, including textbooks,
reviews, tutorials on this topic
\citep{bishop2006pattern,Murphy:2012:MLP:2380985, blei2017variational,
    zhang2018advances} where natural-gradients are often put in a
special category. 
\updated{Our work fixes this issue.}

We will show that natural gradients retrieve essential higher-order
information about the loss landscape which are then assigned to 
appropriate natural parameters using \cref{eq:fixed_pt_specific}. The
information-matching is due to the presence of the entropy term there,
which is an important quantity for the
optimality of Bayes in general \citep{jaynes1982, art669} and 
generally absent in non-Bayesian formulations, for instance, \cref{eq:ERM}.
The entropy term in general leads to exponential-weighting in 
Bayes' rule \citep{littlestone1994weighted, Vovk:1990:AS:92571.92672}.
In our context, it gives rise to natural-gradients and,  
as we will soon see, automatically determines the complexity of the
derived algorithm through the complexity of the class of distributions
$\mathcal{Q}$, yielding a principled way to develop new algorithms.

Overall, our work demonstrates the importance of natural-gradients
and information geometry for algorithm design in ML. This
is similar in spirit to Information Geometric Optimization
\citep{ollivier2017information} which focuses on the optimization of
black-box, deterministic functions. In contrast, we derive generic
learning algorithms by using a Bayesian \updated{objective where an additional entropy term is added}.

The BLR we
use is a generalization of the method \updated{by} 
\citet{khan2017conjugate, khan2018fast1}, \updated{proposed} specifically for
\updated{variational} inference. Here, we
establish it as a general learning rule to derive many old and new
learning algorithms. These include both Bayesian and
    non-Bayesian algorithms, \updated{going} way beyond its original proposal. 
    We do not claim that these successful algorithms work well because
    they are derived from the BLR. Rather, we use the BLR to simply
    unravel the inherent Bayesian nature of these ``good''
    algorithms. In this sense, the BLR can be seen as a variant of
Bayes' rule, useful for generic algorithm design.

\subsection{Examples}
\label{sec:examples}

To fix ideas, we will now use the BLR to derive three classical
algorithms: gradient descent, Newton's method, and ridge regression.
\updated{Such derivations are repeated throughout the paper to recover various algorithms in different fields. A full list of the learning algorithms derived in this paper is given
in \cref{tab:summary}. We not only unify and generalize existing algorithms but also derive new ones. These includes: (i) a new multimodal optimization algorithm, (ii) new uncertainty estimation algorithms (OGN and VOGN), (iii) the BayesBiNN algorithm for binary neural networks, and (iv) non-conjugate variational inference algorithms.

For notational simplicity, we will write $q_{\vnatparam_t}$ as $q_t$ in the rest of the paper. Throughout this section, we choose $\mathcal{Q}$ to be the set of multivariate Gaussians and use the following simplified form of the BLR,
\begin{equation}
  \vnatparam_{t+1} \leftarrow (1- \rho_t) \vnatparam_t -
  \rho_t \grad_{\vmeanparam}
  \myexpect_{q_t} \sqr{\barloss(\vparam) + \log h(\vparam)},
  \label{eq:bayes_learn_rule}
\end{equation}
This is obtained by using $\grad_{\vmeanparam} \mathcal{H}(q) = - \vnatparam -
\grad_{\vmeanparam} \myexpect_q(\log h(\vparam))$; see the derivation in \cref{app:natgrad_entropy}.}

\subsubsection{Gradient Descent}
\label{sec:grad_desc}

The gradient descent algorithm uses the following update
\begin{displaymath}
    \vparam_{t+1} \leftarrow \vparam_t -
    \rho_t \grad_{\text{\vparam}} \barloss(\vparam_t),
\end{displaymath}
using only the first-order information
$\grad_{\text{\vparam}} \barloss(\vparam_t)$ (the gradient evaluated
at $\vparam_t$). We choose $q(\vparam) = \gauss(\vparam|\vm, \vI)$, a
multivariate Gaussian with unknown mean $\vm$ and known covariance
matrix set to $\vI$ for simplicity; a more general case is given in \cref{eq:sgd_cov}.

%
\updated{We can now simply plug in the definition of $\vnatparam, \vmeanparam,$ and $h(\vparam)$ in \cref{eq:bayes_learn_rule}. For $q(\vparam) = \gauss(\vparam|\vm, \vI)$, we have $\vnatparam=\vmeanparam=\vm$ (\cref{tab:EFsummary}).} We also have $\log h(\vparam) = -\half P\log(2\pi) -\half\vparam^{T}\vparam$.
\updated{Using these in \cref{eq:bayes_learn_rule}, the update} follows directly:
\begin{align}
  \vm_{t+1} \leftarrow \vm_t - \rho_t \left. \grad_{\text{\vm}}
   \myexpect_{\text{\gauss}(\text{\vparam}|\text{\vm},\vI)} \sqr{\barloss(\vparam)} \right\vert_{\text{\vm} = \text{\vm}_t}.
  \label{eq:blr_sgd1}
\end{align}
This is gradient descent but over the expected loss
$\myexpect_{q} \sqr{\barloss(\vparam)}$. We can remove the expectation
by using the first-order delta method \citep{dorfman1938note,
    ver2012invented} (\cref{app:delta}) \updated{where we approximate the expectation of a function by its value at the mean,}
\begin{displaymath}
   \grad_{\vm} \myexpect_{q} \sqr{ \barloss(\vparam)} 
   \approx \left.
  \grad_{\vparam} \barloss(\vparam)\right\vert_{\vparam =
    \text{\vm}}.
\end{displaymath}
With this, the BLR equals the gradient descent algorithm with $\vm_t$
as the iterate $\vtheta_t$.  The two choices, first of the distribution
$\mathcal{Q}$ and second of the delta method, give us gradient descent
from the BLR.

\updated{The delta method allows us to interpret non-Bayesian solutions as crude approximations of Bayesian ones. By approximating the 
    expectation} in the BLR with a greedy approximation, we are back to
    minimizing a deterministic (non-Bayesian) objective. This
    observation suggests that \cref{eq:blr_sgd1} \updated{should perform better than gradient descent. In \cref{sec:uncertainty_dl}, we will revisit this point and see that averaging in fact leads to improved uncertainty and better robustness properties.}
Using the Dirac's-delta distribution instead of the delta
    method is not advisable, as their use in this context is
    fundamentally flawed. This is because the entropy goes to $-\infty$, 
    making the Bayesian objective meaningless
    \citep{welling2008deterministic}. 

\subsubsection{Newton's Method}
\label{sec:blr_newton}

Newton's method is a second-order method
\begin{align}
  \vparam_{t+1} \leftarrow \vparam_t -
  \sqr{\grad_{\text{\vparam}}^2 
  \barloss(\vparam_t)}^{-1} \sqr{ \grad_{\text{\vparam}} 
  \barloss(\vparam_t)}.
\end{align}
which too can be derived from the BLR by expanding the class
$\mathcal{Q}$ to $q(\vparam) = \gauss(\vparam|\vm, \vS^{-1})$ with an
unknown precision matrix $\vS$. This example illustrates a property of
the BLR: the complexity of the derived algorithm is directly related
to the complexity of the exponential family ${\mathcal Q}$. Fixed
covariances previously gave rise to gradient-descent (a first-order
method) and by increasing the complexity where we also learn the
precision matrix, the BLR reduces to a (more complex) second-order
method.

We use the simplified form of the BLR given in \cref{eq:bayes_learn_rule}. The function $h(\vparam)$ is a
constant, and the natural and expectation parameters divide themselves
into natural pairs
\begin{equation}
  \begin{array}{llll}
    \begin{array}{cc}
      \vnatparam^{(1)} &= \vS\vm, \\
      \vnatparam^{(2)} &=-\half \vS,
    \end{array}
                       &\quad
                         \begin{array}{lll} 
                           \vmeanparam^{(1)} & = \myexpect_{q}[\vparam] &= \vm, \\
                           \vmeanparam^{(2)} & =
                                               \myexpect_{q}[\vparam\vparam^\top]
                                                                        &= \vS^{-1} + \vm\vm^T. \\
                         \end{array}
  \end{array}
  \label{eq:full_gauss_params}
\end{equation}
The natural gradients can be expressed in terms of the gradient and
Hessian of $\barloss(\vtheta)$.
\updated{To show this, we use the chain-rule to first write gradients with respect to $\vmu$ in terms of $(\vm,\vS^{-1})$ (first equality below) and then use Bonnet's and Price's theorem respectively
\citep{bonnet1964transformations, price1958useful,
rezende2014stochastic} to get the second equality (full derivation is in \cref{app:bp}),}
\begin{align}
  \grad_{\vmeanparam^{(1)}} \myexpect_{q}
  [\barloss(\vparam)]
  &= \grad_{\boldsymbol{\mathbf{m}}} \myexpect_{q}
    [\barloss(\vparam)]
    - 2 \sqr{ \grad_{\mathbf{S}^{-1}} \myexpect_{q}
    [\barloss(\vparam)] } \vm
  &&= \myexpect_{q} [ \grad_{\boldsymbol{\param}}
     \barloss(\vparam)] - \myexpect_{q} [
     \grad_{\boldsymbol{\param}}^2 \barloss(\vparam)] \vm,
     \label{eq:ngrad_1_gauss}\\
  \grad_{\vmeanparam^{(2)}} \myexpect_{q}
  [ \barloss(\vparam)]
  &= \grad_{\mathbf{S}^{-1}} \myexpect_{q} [
    \barloss(\vparam)]
  &&= \half \myexpect_{q} [ \grad_{\boldsymbol{\param}}^2 \barloss(\vparam)].
     \label{eq:ngrad_2_gauss}
\end{align}
The expressions show that 
the natural gradients contain the information about the first and
second-order derivatives of $\barloss(\vparam)$.

The BLR now turns into an online variant of Newton's method where the
precision matrix contains an exponential-smoothed Hessian average, 
used as a pre-conditioner to update the mean,
\begin{align}
  \vm_{t+1} \leftarrow \vm_t -
  \rho_t \vS_{t+1}^{-1} \myexpect_{q_t}
  \sqr{ \grad_{\text{\vparam}}\barloss(\vparam)} 
  \quad\text{and}\quad
  \vS_{t+1} \leftarrow (1-\rho_t) \vS_t +
  \rho_t \myexpect_{q_t} \sqr{
  \grad_{\text{\vparam}}^2 
  \barloss(\vparam)}.
  \label{eq:von_1}
\end{align}
We can recover the classical Newton's method in three steps. First,
apply the delta method (see \updated{\cref{eq:firstderiv,eq:secondderiv} in} \cref{app:delta}),
\begin{equation}
   \myexpect_{q} \sqr{ \grad_{\text{\vparam}} \barloss(\vparam)}
    \approx \left. \grad_{\vparam} \barloss(\vparam)\right\vert_{\vparam =
    \vm} 
      \quad\text{and}\quad
   \myexpect_{q} \sqr{ \grad_{\text{\vparam}}^2 \barloss(\vparam)}
    \approx \left. \grad^2_{\vparam} \barloss(\vparam)
   \right\vert_{\vparam = \vm}. 
   \label{eq:bp_delta}
\end{equation}
Second, set the learning rate to 1 which is justified when the loss is
strongly convex or the algorithm is initialized close to the
solution. Finally, treat the mean $\vm_t$ as the iterate $\vparam_t$.

\subsubsection{Ridge Regression}
\label{sec:ridge}

Why do we get a second-order method when we increase the complexity of
the Gaussian? This is due to the natural gradients which, depending on
the complexity of the distribution, retrieve essential higher-order
information about the loss landscape. We illustrate this now through
the simplest non-trivial case of Ridge regression. \updated{Denote input matrix by $\vX\in\real^{N\times D}$ and output by $\vy\in\real^N$.} The loss is
quadratic:
$\barloss(\vparam) = \half (\vy - \vX \vparam)^\top (\vy - \vX\vparam)
+ \half \delta \vparam^\top \vparam$ with $\delta>0$ as the
regularization parameter, and the solution is available in
closed-form:
\begin{displaymath}
    \vparam_* = (\vX^\top \vX + \delta\vI)^{-1}\vX^\top \vy .
\end{displaymath}
\updated{Natural gradients for this case are extremely easy to derive.} 
We first note that the expected-loss is linear in $\vmeanparam$, \updated{that is,}
\[\myexpect_{q} \sqr{\bar{\loss}(\vparam)} = - \vy^\top \vX
\vmeanparam^{(1)} + \trace \sqr{ \half \rnd{\vX^T \vX + \delta\vI }
    \vmeanparam^{(2)} } + \half \vy^\top \vy,\] and therefore the natural-gradients are \updated{simply the coefficients in front of $\vmu^{(1)}$ and $\vmu^{(2)}$:}
\begin{equation}
  \grad_{\vmeanparam^{(1)}}
  \myexpect_{q} \sqr{\bar{\loss}(\vparam) } =  -
  \vX^T \vy
  \quad\text{and}\quad
  \grad_{\vmeanparam^{(2)}}
  \myexpect_{q} \sqr{\bar{\loss}(\vparam) } =
  \half \rnd{\vX^T \vX + \delta\vI }.
  \label{eq:natgrad_ridge}
\end{equation}
\updated{These already contain parts of the solution $\vparam_*$
which we can recover by using \cref{eq:fixed_pt_specific}}
\begin{displaymath}
    \vnatparam_* = \grad_{\vmeanparam} \myexpect_{q_{\vnatparam_*}}\sqr{-\barloss(\vparam)} \,\, \implies \,\,
  \vS_*\vm_* =  \vX^T \vy, \quad\quad
  \vS_* = \vX^T \vX + \delta\vI .
\end{displaymath}
and solving to get the mean $\vm_* = \vparam_*$. By increasing the
complexity of $\mathcal{Q}$, natural-gradients can retrieve
appropriate higher-order information, which are then assigned to the
corresponding natural parameters using \cref{eq:fixed_pt_specific}.
This is the main reason why different algorithms are obtained when we
change the class $\mathcal{Q}$. We discuss this point further in
\cref{sec:opt} when relating Bayesian principles to those of
optimization.

\renewcommand{\arraystretch}{1.3}
\begin{table}[!ht]
         \begin{center}
        \begin{tabular}{p{1.5in} p{1.4in} p{2.3in} p{.2in}}
          \toprule
          {\bf Learning Algorithm} & {\bf Posterior Approx.} & {\bf Natural-Gradient Approx.} & {\bf Sec.}\\
          \bottomrule
          \rowcolor{Gainsboro!60}
          \multicolumn{4}{c} {\bf Optimization Algorithms}\\
          Gradient Descent & Gaussian (fixed cov.) & Delta method & \ref{sec:grad_desc} \\
          Newton's method & Gaussian & -----``----- & \ref{sec:blr_newton} \\
          Multimodal optimization {\tiny (New)} & Mixture of Gaussians & -----``----- & \ref{sec:multimodal} \\
          \bottomrule
          \rowcolor{Gainsboro!60}
          \multicolumn{4}{c} {\bf Deep-Learning Algorithms}\\
          SGD & Gaussian (fixed cov.)  & Delta method, stochastic approx. & \ref{sec:sgd} \\
          RMSprop/Adam & Gaussian (diagonal cov.) & Delta method, stochastic approx., Hessian approx., square-root scaling, slow-moving scale vectors & \ref{sec:ada_dl} \\
          Dropout & Mixture of Gaussians & Delta method, stochastic approx., responsibility approx. & \ref{sec:dropout} \\
          STE & Bernoulli & Delta method, stochastic approx. & \ref{sec:binn}\\
          Online Gauss-Newton (OGN) {\tiny (New)}& Gaussian~(diagonal cov.) & Gauss-Newton Hessian approx. in Adam \& no square-root scaling & \ref{sec:uncertainty_dl}\\
          Variational OGN {\tiny (New)} & -----``----- & Remove delta method from OGN & \ref{sec:uncertainty_dl}\\
          BayesBiNN {\tiny (New)} & Bernoulli & Remove delta method from STE & \ref{sec:binn}\\
          \bottomrule
          \rowcolor{Gainsboro!60}
          \multicolumn{4}{c} {\bf Approximate Bayesian Inference Algorithms}\\
          Conjugate Bayes & Exponential family & Set learning rate $\rho_t= 1$ & \ref{sec:conj_bayes} \\
          Laplace's method & Gaussian & Delta method & \ref{sec:uncertainty_dl} \\
          Expectation-Maximization & Exponential Family + Gaussian & Delta method for the parameters & \ref{sec:em} \\
          Stochastic Variational Inference (SVI) & Exponential family (mean-field) & Stochastic approx., local $\rho_t= 1$   & \ref{sec:svi} \\
           Variational Message Passing (VMP) & -----``----- & $\rho_t= 1$ for all nodes & \ref{sec:svi} \\
          Non-Conjugate VMP & -----``----- & -----``----- & \ref{sec:svi} \\
          Non-Conjugate Variational Inference {\tiny (New)} & Mixture of Exponential family  & None & \ref{sec:ncvi} \\
          \bottomrule
        \end{tabular}
      \end{center}
      \caption{A summary of learning algorithms derived from the BLR,
        along with the required approximations to the posterior 
        and natural gradients.  New algorithms are 
        marked with ``New''.  Abbreviations: cov. $\to$ covariance, STE $\to$ Straight-Through-Estimator.
      }
      \label{tab:summary}
    \end{table}

\subsection{Outline of the Rest of the Paper}

The rest of the paper is organized as follows.
In \cref{sec:bayes_learn}, we give two derivations of the BLR by using
natural-gradient descent and mirror descent, respectively. In
\cref{sec:opt}, we summarize the main principles for the derivation of
optimization algorithms, and give guidelines
for the design of new multimodal-optimization algorithms. In
\cref{sec:dl}, we discuss the derivation of existing deep-learning
algorithms, and design new algorithms for uncertainty estimation. In
\cref{sec:approx_bayes}, we discuss the derivation of algorithms for
Bayesian inference for both conjugate and non-conjugate models. In
\cref{sec:discussion}, we conclude.

\section{Derivation of the Bayesian Learning Rule}
\label{sec:bayes_learn}

This section contains two derivations of the BLR. First, we interpret
it as a natural-gradient descent using a second-order expansion of the
KLD, which strengthens its intuition. Second, we do a more formal
derivation \updated{specifically for exponential-family, where we use} a mirror-descent algorithm leveraging the connection
to Legendre-duality. \updated{This derivation is more direct and obtained
without the second-order approximation. It reveals that
certain parameterizations are preferable for natural-gradient
descent over Bayesian objectives.}

\subsection{Bayesian Learning Rule as Natural-Gradient Descent}
\label{sec:natgrad}

\updated{By expanding the KLD term and collecting the log-prior term with the loss functions, we can write the objective in \cref{eq:BayesP} as follows,
\[
   \mathcal{L}(\vnatparam) = \myexpect_{q} [\barloss(\vparam) + \log
q(\vparam)].
\]
}The classical gradient-descent
algorithm to minimize $\mathcal{L}(\vnatparam)$ performs the following update,
\begin{align}
  \vnatparam_{t+1} \leftarrow \vnatparam_t -
  \rho_t \grad_{\boldsymbol{\natparam}} \mathcal{L}(\vnatparam_t).
   \label{eq:bayes_gd}
\end{align}
The insight motivating natural-gradient algorithms, is that 
\cref{eq:bayes_gd} solves the following:
\begin{align}
  \vnatparam_{t+1} \leftarrow \arg\min_{\boldsymbol{\natparam}} \,\,
  \myang{ \grad_{\boldsymbol{\natparam}} \mathcal{L}(\vnatparam_t), \vnatparam
  } +
  \frac{1}{2\rho_t} \| \vnatparam - \vnatparam_t \|_2^2,
   \label{eq:bayes_gd_euclid}
\end{align}
which reveals the implicit Euclidean penalty of changes in the parameters.
The parameters $\vnatparam_t$ parameterize probability
distributions, and therefore their updates should be penalized based
on the distance in the space of distributions.
Distance between two
parameter configurations might be a poor measure of the distance
between the corresponding distributions.
Natural-gradient algorithms~\citep{amari1998natural, martens2014new}
use instead an alternative penalty, and using the KLD, 
we get
\begin{align}
   \vnatparam_{t+1} \leftarrow \arg\min_{\boldsymbol{\natparam}}
  \,\,
  \myang{ \grad_{\boldsymbol{\natparam}} \mathcal{L}(\vnatparam_t) ,
  \vnatparam} +
  \frac{1}{\rho_t} \dkls{}{q(\vparam)}{q_t(\vparam)}.  
   \label{eq:bayes_ngd_distance}
\end{align}
\updated{A closed form update can be obtained with a second-order expansion of the
KLD-term which is obtained by using the fact that the Hessian of the KLD with respect to
    $\vnatparam$ is equal to the FIM $\vF(\vnatparam)$ of $q(\vparam)$~\citep{pascanu2013revisiting}, that is,
    \[
       \dkls{}{q(\vparam)}{q_t(\vparam)} \approx \half (\vnatparam - \vnatparam_t)^\top \vF(\vnatparam_t) (\vnatparam - \vnatparam_t)
    \]
    Using this, we get}
\begin{equation}
      \vnatparam_{t+1} 
      \leftarrow \vnatparam_t -
        \rho_t  \vF(\vlambda_t)^{-1} \,\,
        \grad_{\boldsymbol{\natparam}} \mathcal{L}(\vnatparam_t).
   \label{eq:bayes_ngd}
\end{equation}
The descent direction
$\vF(\vlambda_t)^{-1} \,\, \grad_{\boldsymbol{\natparam}_t}
\mathcal{L}(\vnatparam_t)$ in this case is referred to as the
natural-gradient. \updated{When $\vnatparam_t$ is the natural parameter of $q_t(\vparam)$, the above coincides with the BLR update
\cref{eq:bayes_learn_rule_general}}.

\updated{We stress that our use of natural-gradient method is more general than the more popular use for maximum-likelihood estimation~\citep{amari1998natural, martens2014new}. There, a log-likelihood of the form $\log p(\vy; \vparam)$ is maximized by using the fisher $\vF(\vparam)$ of $p(\vy; \vparam)$, 
\begin{equation}
   \vparam_{t+1} \leftarrow \vparam_t - \rho_t \vF(\vparam_t)^{-1} \nabla_\vparam [- \log p(\vy;\vparam)].
\end{equation}
This update uses the same distribution $p(\vy; \vparam)$ to define both the objective and the Fisher, which is in contrast to \cref{eq:bayes_learn_rule_general} where we are free to use any loss $\barloss(\vparam)$ in the objective and the distribution $q(\vparam)$ can also be chosen freely, irrespective of the objective. The BLR update is more general and the above maximum-likelihood update can be obtained as a special instance by using $q$ as a Gaussian distribution, as shown in 
\cref{sec:uncertainty_dl}.}


\subsection{Bayesian Learning Rule as Mirror-Descent}
\label{sec:blr_mirror}

\updated{The previous derivation holds for a general $q(\vparam)$. 
In this section, we will do a derivation specifically for exponential-family distribution by using} a mirror-descent
formulation, similarly to \citet{raskutti2015information}, but for a
Bayesian objective. A mirror-descent step defined in the
space of $\vmeanparam$ \updated{is shown to be equivalent} to a natural-gradient step in the space of
$\vnatparam$ (the update \cref{eq:bayes_ngd}). This is a consequence
of the Legendre-duality between the spaces of $\vnatparam$ and
$\vmeanparam$ \citep{WainwrightJordan08}, and is also related to the dual-flat Riemannian structures
\citep{amari2016information} employed in information geometry. 
Unlike the previous derivation, this derivation is more direct and obtained
without any second-order approximation of the KLD, and it reveals that, due to computational reasons,
certain parameterizations should be preferred for a natural-gradient
descent over Bayesian objectives.

There is a duality between the space of $\vnatparam$ and $\vmeanparam$,
since $\vmeanparam = \grad_{\boldsymbol{\natparam}} A(\vnatparam)$ is
a bijection. An alternative view of this mapping, is to use the
Legendre transform of $A(\vnatparam)$,
\begin{displaymath}
    A^*(\vmeanparam) = \sup_{\vnatparam'\in\Omega}
    \myang{\vmeanparam,\vnatparam'} - A(\vnatparam') .
\end{displaymath}
The mapping follows by setting the gradient of the
right-hand-side to zero.
The reverse mapping is
$A(\vnatparam) = \sup_{\boldsymbol{\meanparam}'\in\mathcal{M}}
{\myang{\vmeanparam',\vnatparam} - A^*(\vmeanparam') }$, giving us,
$\vnatparam = \grad_{\boldsymbol{\meanparam}} A^*(\vmeanparam)$.

The expectation parameters $\vmeanparam$ provide a dual
coordinate system to specify the exponential family with natural
parameter $\vnatparam$, and using $\vmeanparam$ we express
$q(\vparam)$ as
given below (see \citet{banerjee2005clustering} for details):
\begin{align}
  q(\vparam) = h(\vparam) \exp\sqr{
  -\mathbb{D}_{A^*}(\vT(\vparam) \| \vmeanparam) + A^*(\vT(\vparam)) }
\end{align}
where
$\mathbb{D}_{A^*}(\vmeanparam_1 \| \vmeanparam_2) = A^*(\vmeanparam_1)
- A^*(\vmeanparam_2) - \myang{\vmeanparam_1 - \vmeanparam_2,
    \grad_{\boldsymbol{\meanparam}}A^*(\vmeanparam_2)}$ is the Bregman
divergence defined using the function $A^*(\vmeanparam)$. The
Bregman and KL divergences are related, 
\begin{equation}
    \dkls{}{q_1(\vparam)}{ q_2(\vparam)} =
    \mathbb{D}_{A^*}(\vmeanparam_1 \| \vmeanparam_2 ) =
    \mathbb{D}_{A}(\vnatparam_2 \| \vnatparam_1 ).
\end{equation}
The Bregman divergence is equal to the KL divergence between the
corresponding distributions, which can also be measured in the dual
space using the (swapped order of the) natural parameters. We can then
use these Bregman divergences to measure the distances in the two dual
spaces. 

The relationship with the KLD enables us to express natural-gradient
descent in one space as a mirror descent in the dual space. 
Specifically, the following mirror-descent update in the space
$\mathcal{M}$ is equivalent to the natural-gradient descent
\cref{eq:bayes_ngd} in the space of $\Omega$,
\begin{align}
   \vmeanparam_{t+1} \leftarrow \arg\min_{\boldsymbol{\meanparam} \in
  \mathcal{M}}\,\,
   \myang{ \grad_{\boldsymbol{\meanparam}} \tilde{\mathcal{L}}(\vmeanparam_t) ,
  \vmeanparam}
  + \frac{1}{\rho_t} \mathbb{D}_{A^*}(\vmeanparam \| \vmeanparam_t).  
   \label{eq:md_mu}
\end{align}
\updated{The $\tilde{\mathcal{L}}(\vmeanparam) = \mathcal{L}(\vnatparam)$ here is a reparameterized objetive expressed in terms of $\vmeanparam$.}
The proof is straightforward: from the definition for the Bregman
divergence, we find that
\begin{equation}
\grad_{\boldsymbol{\meanparam}} \mathbb{D}_{A^*}(\vmeanparam \|
\vmeanparam_t) = \vnatparam - \vnatparam_t,
\end{equation}
and if we take the derivative of \cref{eq:md_mu} with respect to
$\vmeanparam$ and set it to zero, we get the BLR update.
\updated{The derivation shows that the second-order approximation in the previous derivation (used to obtain \cref{eq:bayes_ngd}) is in fact exact for exponential-family distribution. Due to this we get an equivalence between the natural gradient descent and mirror descent.}

The reverse statement also holds: a mirror descent update in the space
$\Omega$ leads to a natural gradient descent in the space of
$\mathcal{M}$. From a Bayesian viewpoint, \cref{eq:md_mu} is closer to
Bayes' rule since an addition in the natural-parameter space $\Omega$
is equivalent to multiplication in the space of $\mathcal{Q}$ (see
\cref{sec:conj_bayes}). Natural parameters additionally encode conditional
independence (e.g., for Gaussians) which is preferable from a
computational viewpoint (see \citet{malago2015information} for a
similar suggestion). In any case, a reverse version of BLR can always
be used if the need arises.



\section{Optimization Algorithms from the Bayesian Learning Rule}
\label{sec:opt}

\subsection{First- and Second-Order Methods by using Gaussian Candidates}
\label{sec:12order}

The examples in \cref{sec:examples} demonstrate the key ideas behind
our recipe to derive learning algorithms:
\begin{enumerate}
\item Natural-gradients retrieve essential higher-order information
    about the loss landscape and assign them to the appropriate
    natural parameters,
\item Both of these choices are dictated by the entropy of the chosen
    class $\mathcal{Q}$, which automatically determines the complexity
    of the derived algorithms.
\end{enumerate}
These are concisely summarized in the optimality condition of the
Bayes objective (derived in \cref{eq:fixed_pt_specific}).
\begin{align}
  \grad_{\vmeanparam} \myexpect_{q_*} [ \barloss (\vparam)]
  = \grad_{\vmeanparam} \entropy \rnd{q_*} 
  \label{eq:fixed_pt}
\end{align}
From this, we can derive as a special case the optimality condition of
a non-Bayesian problem over $\barloss(\vparam)$, given in \cref{eq:ERM}.
For instance, for Gaussians $\gauss(\vparam|\vm,\vI)$, the entropy is
constant. Therefore the right hand side in \cref{eq:fixed_pt} goes to
zero, giving the first equality below,
\begin{equation}
    \label{eq:conv_cond_1}
    \left. \grad_{\mathbf{m}} \myexpect_{q_*}\sqr{  
          \barloss(\vparam) }\right\vert_{\vm = \vm_*} = 0 
   \xrightarrow[]{\text{Bonnet's thm.}}
          \myexpect_{q_*}\sqr{ \grad_{\text{\vparam}} \barloss(\vparam) } 
          = 0
    \xrightarrow[]{\text{delta}}
    \left. \grad_{\boldsymbol{\param}} \barloss(\vtheta)
    \right\vert_{\vtheta = \vparam_*} = 0.
\end{equation}
The simplifications shown at the right above are
respectively obtained by using Bonnet's theorem (\cref{app:bp}) and
the delta method (\cref{app:delta}) to finally recover the
$1^{\text{st}}$-order optimality condition at a local minimizer
$\vparam_*$ of $\barloss(\vparam)$. Clearly, the above natural
gradient contain the information about the first-order derivative of
the loss, to give us an equation that determines the value of the
natural parameter $\vm_*$.

When the complexity of the Gaussian is increased to include the
covariance parameters with candidates $\gauss(\vparam|\vm,\vS^{-1})$,
the entropy is no more a constant but depends on $\vS$: we have
$\mathcal{H}(q) = -\half\log|2\pi e \vS|$. The fixed-point of the BLR
(in \cref{eq:von_1} now) to yield an additional condition, shown on
the left,
\begin{equation}
    \label{eq:conv_cond_2}
    \left. \grad_{\text{\vS}^{-1}} \myexpect_{q_*}\sqr{ \barloss(\vparam)
          } \right\vert_{\vS^{-1} = \vS_*^{-1}}  = \half \vS_*
    \,\xrightarrow[]{\text{Price's thm.}} \,
   \myexpect_{q_*}\sqr{ \nabla_{\text{\vparam}}^2 \barloss(\vparam)
          } = \vS_*
    \, \xrightarrow[]{\text{delta}} \,
    \left. \grad_{\boldsymbol{\param}}^2
      \barloss(\vtheta) \right\vert_{\vtheta = \vparam_*}\succ 0. 
\end{equation}
The simplifications at the right are respectively obtained by using
Price's theorem (\cref{app:bp}) and the delta method
(\cref{app:delta}) to recover the 2$^\text{nd}$-order optimality
condition of a local minimizer $\vparam_*$ of $\barloss(\vparam)$,
Here, $\succ 0$ denotes the positive-definite condition which follows
from $\vS_*\succ 0$. The condition above matches the second-order
derivatives to the precision matrix $\vS_*$. In general, more complex
sufficient statistics in $q(\vparam)$ can enable us to retrieve
higher-order derivatives of the loss through natural gradients.


\subsection{Multimodal Optimization by using Mixtures-Candidate Distributions}
\label{sec:multimodal}

It is natural to expect that by further increasing the complexity of the class
$\mathcal{Q}$, we can obtain algorithms that go beyond standard
Newton-like optimizers. We illustrate this point by using mixture
distribution to obtain a new algorithm for multimodal optimization
\citep{yu2010introduction}. The goal is to simultaneously locate
multiple local minima within a single run of the algorithm
\citep{wong2015evolutionary}. Mixture distributions are ideal for this
task where individual components can be tasked to locate different
local minima and diversity among them is encouraged by the entropy,
forcing them to take responsibility of different regions. \updated{Our goal in this section is to 
illustrate this principle. Note that we do not aim to propose} new algorithms that
solve multimodal optimization in its entirety.

We will rely on the work of \cite{linfast, lin2019stein} who derive a
natural-gradient update, similar to
\cref{eq:bayes_learn_rule_general}, but for mixtures. Consider, for
example, the finite mixture of Gaussians
\begin{displaymath}
    q(\vparam) = \sum_{k=1}^K \pi_k \gauss(\vparam|\vm_k, \vS_k^{-1})
\end{displaymath}
with $\vm_k$ and $\vS_k$ as the mean and precision of the $k$'th
Gaussian component and $\pi_k$ are the component probabilities with
$\pi_K = 1 - \sum_{k=1}^{K-1} \pi_k$. In general, the FIM of such
mixtures could be singular, making it difficult to use
natural-gradient updates. However, the joint
$q(\vparam,z=k) = \pi_k \gauss(\vparam|\vm_k, \vS_k^{-1})$, where
$z\in\{1,2,\ldots,K\}$ a mixture-component indicator, has a
well-defined FIM which can be used instead. We will now briefly
describe this for the mixture of Gaussian case with fixed $\pi_k$.

We start by writing the natural and expectation parameters of the
joint $q(\vparam,z=k)$, denoted by $\vnatparam_k$ and $\vmeanparam_k$
respectively. Since $\pi_k$ are fixed, these take very similar form to
a single Gaussian case in
\cref{eq:full_gauss_params} \updated{as shown below (we denote the indicator function by $1_{[\lat=k]}$)}, 
\begin{equation}
    \begin{array}{ll}
      \begin{array}{ll} 
        \vnatparam^{(1)}_k &= \vS_k\vm_k, \\
        \vnatparam^{(2)}_k &=-\half \vS_k, 
      \end{array}
       \begin{array}{ll} 
         \qquad\vmeanparam_k^{(1)} & = \myexpect_q
                                    \sqr{ 1_{[\lat=k]} \vparam} = \pi_k\vm_k , \\
         \qquad\vmeanparam_k^{(2)} & = \myexpect_q
                                    \sqr{ 1_{[\lat=k]}
                                    \vparam\vparam^T } =
                                    \pi_k \rnd{ \vS_k^{-1} + \vm_k\vm_k^T }, 
       \end{array}
    \end{array}
    \label{eq:mixgauss_params}
\end{equation}
where $\vnatparam_k = \{\vnatparam_k^{(1)}, \vnatparam_k^{(2)} \}$ and
$\vmeanparam_k = \{\vmeanparam_k^{(1)}, \vmeanparam_k^{(2)}\}$. With
this, we can write a natural-gradient update for
$\vnatparam_k,$ for all $k=1, \ldots, K$, by using the gradient with
respect to $\vmeanparam_k$ \citep[Theorem 3]{linfast},
\begin{equation}
   \vnatparam_{k,t+1} \leftarrow \vnatparam_{k,t} -
       \rho_t \nabla_{\boldsymbol{\meanparam}_k}
       \sqr{ \myexpect_{q_t}
       \rnd{\barloss(\vparam)} - \entropy( q_t) },
       \label{eq:BLRmix}
\end{equation}
where $\vnatparam_{k,t}$ denotes the value of the natural parameter $\vnatparam_k$ at
iteration $t$ and $\entropy(q) = \myexpect_{q} [-\log q(\vparam)]$ is
the entropy of the mixture $q(\vparam)$ (not the joint
$q(\vparam,z)$). This is an extension of the BLR 
(\cref{eq:bayes_learn_rule_general}) to finite mixtures.

As before, the natural gradients retrieve first
and second order information. Specifically, as shown in
\cref{app:mog}, the update for each $k$ takes a Newton-like form,
\begin{align}
   \vm_{k,t+1} 
  & \leftarrow \vm_{k,t} - \rho_t\, \vS_{k,t+1}^{-1}
    \myexpect_{q_{k,t}} \sqr{ \grad_{\text{\vparam}}
    \barloss(\vparam) + \grad_{\text{\vparam}}\log q(\vparam) },
    \label{eq:exp-von1}\\
  \vS_{k,t+1}
  & \leftarrow \vS_{k,t} + \rho_t\, \myexpect_{q_{k,t}}
    \sqr{  \grad_{\text{\vparam}}^2
    \barloss(\vparam) +  \grad_{\text{\vparam}}^2
    \log q(\vparam) },
    \label{eq:exp-von2}
\end{align}
where $q_{k,t}(\vparam) = \gauss(\vparam|\vm_{k,t}, \vS_{k,t}^{-1})$
is the $k$'th component at iteration $t$. The mean and precision of a
component are updated using the expectations of the gradient and
Hessian respectively. Similarly to the Newton step
of \cref{eq:von_1}, the first update is preconditioned with the
covariance. The expectation helps to focus on a region where
the component has a high probability mass. 

The update can locate multiple solutions in a single run, when each
component takes the responsibility of an individual minimum. For
example, for a problem with two local minima $\vparam_{1,*}$ and
$\vparam_{2,*}$, the best candidate with two components
$\gauss(\vparam|\vm_{1,*}, \vS_{1,*}^{-1})$ and
$\gauss(\vparam| \vm_{2,*}, \vS_{2,*}^{-1})$ satisfies optimality
condition,
\begin{align}
   \myexpect_{\gauss(\text{\vparam}|\text{\vm}_{k,*}, \text{\vS}_{k,*}^{-1})} 
   \Big[ \grad_{\text{\vparam}}  \barloss(\vparam) + \underbrace{
   r(\vparam)\,  \vS_{1,*} (\vm_{1,*} - \vparam) + (1-
  r(\vparam))\,
   \vS_{2,*} (\vm_{2,*} - \vparam) }_{= \nabla_{\text{\vparam}} \log q(\text{\vparam})} \Big] = 0,
   \label{eq:mog_opt}
\end{align}
for $k=1,2$, where we use expression for $\grad_{\text{\vparam}}\log q(\vparam)$ 
from \cref{eq:mixexp_1st} in \cref{app:mog}, and
$r(\vparam)$ is the responsibility function
defined as
\begin{equation}
    r(\vparam) = \frac{\pi \gauss(\vparam| \vm_{1,*}, \vS_{1,*}^{-1}) }
    { \pi \gauss(\vparam| \vm_{1,*}, \vS_{1,*}^{-1}) + (1-\pi)
        \gauss(\vparam| \vm_{2,*}, \vS_{2,*}^{-1})}.
        \label{eq:responsibility}
\end{equation}
Suppose that each component takes responsibility of one local minimum
each, for example, for $\vparam$ sampled from the first component,
$r(\vparam) \approx 1$, and for those sampled from the second
component, $r(\vparam) \approx 0$ (see \cref{eq:gaussmix_assumption} in \cref{app:mog} which gives the explicit assumptions). Under this condition, both the
second and third term in \cref{eq:mog_opt} are negligible, and the
mean $\vm_{k,*}$ approximately satisfies the first-order optimality
condition to qualify as a minimizer of $\barloss(\vparam)$,
\begin{equation}
    \myexpect_{\gauss(\text{\vparam}|\text{\vm}_{k,*},
        \text{\vS}_{k,*}^{-1})} [\grad_{\text{\vparam}}
    \barloss(\vparam)] \,\,\, \approx 0  
    \quad \xrightarrow[]{\text{delta}} \quad
    \left. \grad_{\boldsymbol{\param}} \barloss(\vtheta)
    \right\vert_{\vtheta = \vm_{k,*}} \approx 0, \quad \text{ for } k=1,2
    \label{eq:mog_delta}
\end{equation}
This roughly implies that $\vm_{k,*} \approx \vparam_{k,*}$, meaning
the two components have located the two local minima.
The example illustrates the role of entropy in encouraging diversity
through the responsibility function. There remain several practical
difficulties to overcome before we are ensured a good algorithmic
outcome, like setting the correct number of components, appropriate
initialization, and of course degenerate solutions where two
components collapse to be the same. \updated{The example simply illustrates the usefulness of being Bayesian.}

Other mixture distributions can
also be used as shown in \citet{linfast} who consider the following:
finite mixtures of minimal exponential-families, scale mixture of
Gaussians \citep{andrews1974scale, eltoft2006multivariate},
Birnbaum-Saunders distribution \citep{birnbaum1969new}, multivariate
skew-Gaussian distribution \citep{azzalini2005skew}, multivariate
exponentially-modified Gaussian distribution
\citep{grushka1972characterization}, normal inverse-Gaussian
distribution \citep{barndorff1997normal}, and matrix-variate Gaussian
distribution \citep{gupta2018matrix, louizos2016structured}.



\section{Deep-Learning Algorithms from the Bayesian Learning Rule}
\label{sec:dl}

\subsection{Stochastic Gradient Descent}
\label{sec:sgd}

Stochastic gradient descent (SGD) is one of the most popular
deep-learning algorithms due to its computational efficiency. The
computational speedup is obtained by approximating the gradient of the
loss by a stochastic-gradient built using a small subset
${\mathcal M}$ of $M$ examples with $M\ll N$. The subset
${\mathcal M}$, commonly referred to as a mini-batch, is randomly
drawn at each iteration from the full data set using a uniform
distribution, and the resulting mini-batch gradient
\begin{align}
  \widehat{\grad}_{\vparam}
  \barloss(\vparam) = \frac{N}{M} \sum_{i\in\mathcal{M}}
  \grad_{\vparam}
  \loss(y_i, f_{\text{\ensuremath\vparam}}(\vx_i) )  +
  \grad_\vparam R(\vparam),
  \label{eq:mbatch_grad}
\end{align}
is much faster to compute. Similarly to the gradient-descent case in
\cref{sec:grad_desc}, the SGD step can be interpreted as the BLR over
a distribution $q(\vparam) = \gauss(\vparam|\vm,\vI)$ with the unknown
mean $\vm$ and the gradient in \cref{eq:blr_sgd1} being replaced by
the mini-batch gradient. The BLR update is extended to candidates
$q(\vparam) = \gauss(\vparam|\vm,\vS^{-1})$ with any fixed
$\vS\succ \vzero$ to get an SGD-like update where $\vS^{-1}$ is the
preconditioner,
\begin{align}
  \vm_{t+1} \leftarrow \vm_t -
   \rho_t\vS^{-1}\,\left. \widehat{\grad}_{\vparam}
  \barloss(\vparam) \right\vert_{\vparam = \vm_t} .
  \label{eq:sgd_cov}
\end{align}
This is derived similarly to \cref{sec:grad_desc} by using the various quantities
associated with $\gauss(\vparam|\vm,\vI)$, that is, 
$\vnatparam= \vS \vm$, $ \vmeanparam = \vm$, and  
$2\log h(\vparam) = P\log|2\pi\vS|
-\vparam^{T}\vS\vparam$.

The above BLR interpretation sheds light on the ineffectiveness of
first-order methods (like SGD) to estimate posterior approximations.
This is a recent trend in Bayesian deep learning, where a
preconditioned update is preferred due to the an-isotropic SGD noise
\citep{mandt2017stochastic, chaudhari2018stochastic,
maddox2019simple}. Unfortunately, a first-order optimizer (such as
\cref{eq:sgd_cov}) offers no clue about estimating $\vS$. SGD iterates
can be used \citep{mandt2017stochastic, maddox2019simple}, but
iterates and the noise in them are affected by the choice of $\vS$,
which makes the estimation difficult. With our Bayesian scheme, the
preconditioner is estimated automatically using a second-order method
(\cref{eq:von_1}) when we allow $\vS$ as a free parameter. We discuss
this next in the context of adaptive learning-rate algorithms.

\subsection{Adaptive Learning-Rate Algorithms}
\label{sec:ada_dl}

Adaptive learning-rate algorithms, motivated from Newton's method,
adapt the learning rate in SGD with a scaling vector, and we discuss
their relationship to the BLR.~Early adaptive variants relied on the
diagonal of the mini-batch Hessian matrix to reduce the computation
cost~\citep{barto1981goal, becker1988improving}, and use exponential
smoothing to improve stability \citep{10.5555/645754.668382}:
\begin{align} \vparam_{t+1} &\leftarrow \vparam_t - \alpha
\frac{1}{\vs_{t+1}} \circ \widehat{ \grad}_{\vparam}
\barloss(\vparam_t), \quad\text{and} \quad\quad \vs_{t+1} \leftarrow
(1-\beta) \vs_t + \beta \,\, \diag[ \widehat{\grad}_{\vparam}^2
\barloss(\vparam_t)],
                  \label{eq:dhess_update}
\end{align} where $\va\circ\vb$ and $\va/\vb$ denote element-wise
multiplication and division respectively, and the scaling vector,
denoted by $\vs_t$, uses $\diag [ \widehat{\grad}_{\vparam}^2
\barloss(\vparam_t)]$ which denotes a vector whose $j$'th entry is
the mini-batch Hessian
\begin{displaymath} \widehat{\grad}_{\param_j}^2 \barloss(\vparam) =
\frac{N}{M} \sum_{i\in\mathcal{M}} | \grad_{\param_j}^2 \loss(y_i,
\vf_{\mathbf{\param}}(\vx_i) ) | + \grad_{\param_j}^2 R(\vtheta).
\end{displaymath} Note the use of absolute value to make sure that the
entries of $\vs_t$ stay positive (assuming $R(\vparam)$ to be
strongly-convex). The exponential smoothing, a popular tool from the
non-stationary time-series literature \citep{brown1959statistical,
holt1960, gardner1985exponential}, works extremely well for deep
learning and is adopted by many adaptive learning-rate algorithms,
for example, the natural Newton method of \citet{le2010fast} and the vSGD
method by \citet{schaul2013no} both use it to estimate the
second-order information, while more modern variants such as RMSprop
\citep{hintonTieleman}, AdaDelta \citep{zeiler2012adadelta}, and Adam
\citep{kingma2014adam}, use it to estimate the magnitude of the
gradient (a more crude approximation of the second-order information
\citep{bottou2016optimization}).

The BLR naturally gives rise to the update in \cref{eq:dhess_update},
where exponential smoothing arises as a \emph{consequence} and not
something we need to invent. We optimize over candidates of the form
$q(\vparam) = \gauss(\vparam|\vm,\vS^{-1})$ where $\vS$ is constrained
to be a diagonal matrix with a vector $\vs$ as the diagonal. This
choice results in the BLR update shown below (the derivation is identical to
\cref{sec:blr_newton}),
\begin{align} \vparam_{t+1} \leftarrow \vparam_t - \rho_t
\frac{1}{\vs_{t+1}} \circ \grad_{\vparam} \barloss(\vparam_t),
\quad\text{where} \quad \vs_{t+1} \leftarrow (1-\rho_t) \vs_t +
\rho_t\,\, \diag(\grad_{\vparam}^2 \barloss(\vparam_t) ),
  \label{eq:don}
\end{align} to obtain $\vm_t = \vparam_t$ and $\diag(\vS_t) = \vs_t$.
Replacing the gradient and Hessian by their mini-batch approximations
and then employing different learning-rates for $\vparam_t$ and
$\vs_t$, we recover the update in \cref{eq:dhess_update}.
\updated{Using different learning rates is useful to compensate for the errors due to minibatches and diagonal approximation of the Hessian.}

The similarity of the BLR update can be exploited to compute
uncertainty estimates for deep-learning models. \citet{khan2018fast}
study the relationship between \cref{eq:don} and modern adaptive
learning-rate algorithms, such as RMSprop \citep{hintonTieleman},
AdaDelta \citep{zeiler2012adadelta}, and Adam \citep{kingma2014adam}.
These modern variants also use exponential smoothing but over
gradient-magnitudes ${\widehat{ \grad}_{\vparam} \barloss(\vparam_t)
\circ \widehat{ \grad}_{\vparam} \barloss(\vparam_t)}$, instead of the
Hessian. For example, RMSprop uses the following update,
\begin{align} \vparam_{t+1} &\leftarrow \vparam_t - \alpha
\frac{1}{\sqrt{\vv_{t+1}} + c \vone} \circ \sqr{\widehat{
\grad}_{\text{\vparam}} \barloss(\vparam_t) }, \quad\text{where} \,\,
\vv_{t+1} \leftarrow (1-\beta) \vv_t + \beta \sqr{\widehat{
\grad}_{\text{\vparam}} \barloss(\vparam_t) \circ \widehat{
\grad}_{\mathbf{\param}} \barloss(\vparam_t)}.
                  \label{eq:rmsprop_update}
\end{align}
There are also some other minor differences to
\cref{eq:don}, for example, the scaling uses a square-root to take
care of the factor $1/M^2$ in the square of the mini-batch gradient
and a small $c>0$ is added to avoid (near)-zero $\vv_{t+1}$
\citep[Sec.~3.4]{khan2018fast}. Despite these small differences, the
similarly of the two updates in \cref{eq:rmsprop_update,eq:don}
enables us to incorporate Bayesian principles in deep learning.
Uncertainty is then computed using a slightly-modified RMSprop update
\citep{osawa2019practical} which, with just a slight changes in the
code, can exploit all the tricks-of-the-trade of deep learning such as
batch normalization, data augmentation, clever initialization, and
distributed training (also see \cref{sec:uncertainty_dl}). The BLR
update extends the application of adaptive learning-rate algorithms to
uncertainty estimation in deep learning.

\updated{The update in \cref{eq:don} also demonstrate that the BLR can naturally deal with stochastic approximations. The update is an online extension of Newton's method where the scale vector uses exponential smoothing. Such smoothing leads to much more stable than directly using a mini-batch approximation of the Hessian in Newton's method. The BLR can lead to second-order methods that are most suitable for online, stochastic, or incremental learning.}

\updated{The BLR update can also be used to recover the Adam optimizer \citep{kingma2014adam}, which is inarguably one of the
most popular optimizer. The Adam optimizer} is closely related
to a BLR variant which uses a momentum based
mirror-descent. Momentum is a common technique to improve convergence
speed of deep learning~\citep{sutskever2013importance}. The classical
momentum method is based on the Polyak's heavy-ball method
\citep{polyak1964some},
\begin{equation*} \vparam_{t+1} \leftarrow \vparam_t - \alpha
\widehat{ \grad}_{\vparam} \barloss(\vparam_t) + \gamma_t (\vparam_t -
\vparam_{t-1})
\end{equation*} with a fixed momentum coefficient $\gamma_t>0$, but
the momentum used in adaptive learning-rate algorithms, such as Adam,
differ in the use of scaling with $\vs_t$ in front of the momentum
term $\vparam_t -\vparam_{t-1}$ (see \cref{eq:adam_moentum_alternate}
in \cref{sec:dl_momentum}). These variants can be better explained by
including a momentum term in the BLR within the mirror descent
framework of \cref{sec:bayes_learn}, which gives the following
modification of the BLR (see \cref{sec:dl_momentum}):
\begin{equation*} \vnatparam_{t+1} \leftarrow \vnatparam_t - \rho_t
\natgrad_\vnatparam \sqr{ \myexpect_{q_t} \rnd{\barloss(\vparam)} -
\entropy(q_t) } + \gamma_t (\vnatparam_t - \vnatparam_{t-1}).
\end{equation*} For a Gaussian distribution, this recovers both SGD
and Newton's method with momentum. Variants like Adam can then be
obtained by making approximations to the Hessian and by using a
square-root scaling; see \cref{eq:don_momentum_1} in
\cref{sec:dl_momentum}.

Finally, we will also mention \citet{aitchison2018bayesian} who derive
deep-learning optimizers using a Bayesian filtering approach with
updating equations similar to the ones we presented. They further
discuss modifications to justify the use of square-root in Adam and
RMSprop. The approach taken by \cite{ollivier2018online} is similar,
but exploits the connection of Kalman filtering and the
natural-gradient method.

\subsection{Dropout}
\label{sec:dropout}

Dropout is a popular regularization technique to avoid overfitting in
deep learning where hidden units are randomly dropped from the neural
network (NN) during training \citep{Srivastava2014}.
\cite{gal2016dropout} interpret SGD-with-dropout as a Bayesian
approximation (called MC-dropout), but their crude approximation to
the entropy term prohibits extensions to more flexible posterior
approximations. In this section, we show that by using the BLR we can
improve their approximation and go beyond SGD to derive MC-dropout
variants of Newton's method, Adam and RMSprop.

Denoting the $j$'th unit in the $l$'th layer by $f_{jl}(\vx)$ for an
input $\vx$, we can write the $i$'th unit for the $l+1$'th layer as
follows:
\begin{align} f_{i,l+1}(\vx) = h \left( \sum_{j=1}^{n_l} \param_{ijl}
z_{jl} f_{jl}(\vx)\right)
\end{align} where the binary weights $z_{jl}\in\{0, 1\}$ are to be
defined, $n_l$ is the number of units in the $l$'th layer, all
$\param_{ijl} \in\real$ are the parameters, and $h(\cdot)$ is a
nonlinear activation function. For simplicity, we have not included an
offset term.

Without dropout, all weights $z_{jl}$ are set to one. With
dropout, all $z_{jl}$'s are independent Bernoulli variables with
probability for being 1 equal to $\pi_1$ (and fixed). If $z_{jl}=0$,
then the $j$'th unit of the $l$'th layer is dropped from the NN, otherwise
it is retained. Let $\tilde{\param}_{ijl}=\param_{ijl}z_{jl}$ and let
$\vparam$ and $\tilde{\vparam}$ denote the complete set of parameters.
The training can be performed with the following simple modification
of SGD where $\tilde{\vparam}$ are used during back-propagation,
\begin{align}
  \vparam_{t+1} &\leftarrow \vparam_t -
                  \alpha \widehat{\grad}_{\vparam}
                  \barloss(\tvparam_t).
                  \label{eq:sgd_dropout}  
\end{align}
This simple procedure has proved to be very effective in practice and
dropout has become a standard trick-of-the-trade to improve
performance of deep networks trained with SGD.

We can include dropout into the Bayesian learning rule, by considering
a spike-and-slab mixture distribution for
$\vparam_{jl}=(\param_{1jl}, \ldots, \param_{n_ljl})$
\begin{align}
  q(\vparam_{jl}) = \pi_1 \gauss(\vparam_{jl}| \vm_{jl},
  \vS_{jl}^{-1}) + (1-\pi_1) \gauss(\vparam_{jl}| \vzero, s_0^{-1}
  \vI_{n_l})
  \label{eq:var_dist_dropout}
\end{align}
where the means and covariances are unknown, but $s_0$ is fixed to a
small positive value to emulate a spike at zero. Further, we assume
independence for every $j$ and $l$,
$q(\vparam)=\prod_{jl} q(\vparam_{jl})$. For this choice,
we arrive in the end at the following update,
\begin{align}
  \vparam_{jl,t+1} &\leftarrow \vparam_{jl,t} - \rho_t\,
                     \vS^{-1}_{jl,t+1} \sqr{ \widehat{\grad}_{\text{\vparam}_{jl}}
                     \barloss(\tvparam_t) } /\pi_1, \label{eq:von_dropout_1}\\
  \vS_{jl,t+1} &\leftarrow \rnd{ 1-\rho_t } \vS_{jl,t} +
                 \rho_t \, \sqr{ \widehat{\grad}_{\text{\vparam}_{jl}}^2
                 \barloss(\tvparam_t) }/\pi_1.\label{eq:von_dropout_2}
\end{align}
The derivation in \cref{app:dropout} uses two approximations. First,
the delta method (\cref{eq:grad_loss}, also used by
\citet{gal2016dropout}), and second, an approximation to the
responsibility function (\cref{eq:responsibility}) which assumes that
$\vparam_{jl}$ is far apart from 0 (can be ensured by choosing $s_0$
to be very small; see \cref{eq:grad_logq,eq:hess_logq}).

The update is a variant of the Newton's method derived in
\cref{eq:von_1}. The difference is that the gradients and Hessians are
evaluated at the dropout weights $\tvparam_t$. By choosing
$\vS_{jl,t}$ to be a diagonal matrix as in \cref{sec:ada_dl}, we can
similarly derive RMSprop/Adam variants with dropout. The BLR
derivation enables more flexible posterior approximations than the
original derivation by \citet{gal2016dropout}.

We can build on the mixture candidate model to learn $\pi_1$ and also 
 allow it to be different across layers, although by doing so we
leave the justification leading to dropout. Such extensions can be
obtained using the extension to mixture distribution by
\citet{linfast}. The update requires gradients with respect to $\pi_1$
which can be approximated by using the Concrete distribution
\citep{maddison2016concrete, jang2016categorical} where the binary
$z_{jl}$ is replaced with an appropriate variable in the unit
interval, see also \cref{sec:binn} for a similar approach.
\cite{gal2017concrete} proposed a similar extension but their approach
do not allow for more flexible candidate distributions. Our approach
here extends to adaptive learning-rate optimizers, and can be extended
to more complex candidate distributions.

\subsection{Uncertainty Estimation in Deep Learning} 
\label{sec:uncertainty_dl}

The observation that some deep-learning algorithms are
    specific instances of the BLR, can be leveraged to improve them. Specifically, we will show that, by improving the
    approximations, we get improved uncertainty estimates. The
    solutions will also have better robustness properties, similar to
    those of flatter minima found by stochastic methods in deep
    learning.

Uncertainty estimation in deep learning is crucial for many
applications, such as medical diagnosis and self-driving cars, but its
computation is extremely challenging due to a large number of
parameters and data examples. A straightforward application of
Bayesian methods does not usually work. Standard Bayesian approaches,
such as Markov Chain Monte Carlo (MCMC), Stochastic Gradient Langevin
Dynamics (SGLD), and Hamiltonian Monte-Carlo (HMC), are either
infeasible or slow~\citep{balan2015bayesian}. Even approximate
inference approaches, such as Laplace's
method~\citep{mackay1991thesis, ritter2018scalable}, variational
inference~\citep{graves2011practical, blundell2015weight}, and
expectation propagation~\citep{hernandez15pbp}, do not match the
performance of the standard point-estimation methods at large scale.
On the ImageNet data set, for example, these Bayesian methods performed
poorly until recently when \citet{osawa2019practical} applied the
natural-gradient algorithms \citep{khan2018fast, zhang2018noisy}. Such
natural-gradient algorithms are in fact variants of the BLR and we
will now discuss their derivation, application, and suitability for
uncertainty estimation in deep learning.

We start with the Laplace's method which is one of the simplest
approaches to obtain uncertainty estimates from point estimates; see
\cite{mackay1991thesis} for an early application to NN. Given a
minimizer $\vparam_*$ of ERM in \cref{eq:ERM} (assume that the loss
correspond to a negative log probability over outputs $y_i$), in
Laplace's method, we employ a Gaussian approximation
$\gauss(\vparam|\vparam_*, \vS_*^{-1})$ where
$\vS_* = \grad_{\text{\vparam}}^2 \barloss(\vparam_*)$. For models
with billions of parameters, even forming $\vS_*$ is infeasible and a
diagonal approximation is often employed \citep{ritter2018scalable}.
Still, Hessian computation after training requires a full pass through
the data which is expensive and challenging. The BLR updates solve this issue
since Hessian can be obtained during training, for example, using
\cref{eq:don} with a minibach gradient and Hessian (denoted by
$\widehat{\grad}$),
\begin{align}
  \vparam_{t+1} \leftarrow \vparam_t -
   \rho_t \frac{1}{\vs_{t+1}} \circ \widehat{\grad}_{\text{\vparam}}
  \barloss(\vparam_t), \quad\text{with} \quad
  \vs_{t+1} \leftarrow (1-\rho_t) \vs_t +
   \rho_t\,\,  \diag(\widehat{\grad}_{\text{\vparam}}^2 
  \barloss(\vparam_t) ).
  \label{eq:don_minibatch}
\end{align}
The vectors $\vs_t$ converge to an unbiased estimate of
the diagonal of the Hessian at $\vparam_*$.

Using (diagonal) Hessian is still problematic in deep learning because
sometimes it can be negative, and the steps can diverge. An additional
issue is that its computation is cumbersome in existing software which
mainly support first-order derivatives only. Due to this, it is
tempting to simply use the RMSprop/Adam update
\cref{eq:rmsprop_update}, but using
$ (\widehat{ \grad}_{\param_j} \barloss(\vparam_t) )^2$ 
results in poor performance as shown by
\citet[Thm.~1]{khan2018fast}. Instead, they propose to use the
following Gauss-Newton approximation which is built from first-order
derivatives but gives better estimate of uncertainty:
\begin{equation}
    \widehat{\grad}_{\param_j}^2 \barloss(\vparam) \approx  
    \frac{N}{M} \sum_{i\in\mathcal{M}} \sqr{ \grad_{\param_j}
    \loss(y_i, f_{\mathbf{\param}}(\vx_i) ) }^2 + \grad_{\param_j}^2 R(\vtheta).
        \label{eq:OGN_approx}
\end{equation}
This is referred to as the online Gauss-Newton
(OGN) in \citet[App.~C]{osawa2019practical}.

The update \cref{eq:don_minibatch} is similar to RMSprop, and can be
implemented with minimal changes to the existing software and many
practical deep-learning tricks can also be applied to boost the
performance. \citet{osawa2019practical} use OGN with batch
normalization, data augmentation, learning-rate scheduling, momentum,
initialization tricks, and distributed training. With these tricks,
OGN gives similar performance as the Adam optimizer, even at large
scale \citep{osawa2019practical} while yielding a reasonable estimate of the
variance. The compute cost is only slightly higher, but it is not an
increase in the complexity, rather due to a difficulty of implementing
\cref{eq:OGN_approx} in the existing deep-learning software.
\citet{dnn2gp} further propose a slightly better (but costlier)
approximation based on the FIM the log-likelihood, to get an algorithm
they call the online Generalized Gauss-Newton (OGGN) algorithm.~Both
OGN and OGGN algorithms are scalable BLR variants for Laplace's method for NNs.

\begin{figure}[t]
    \center
    \subfigure[]{\includegraphics[height=2.5in]{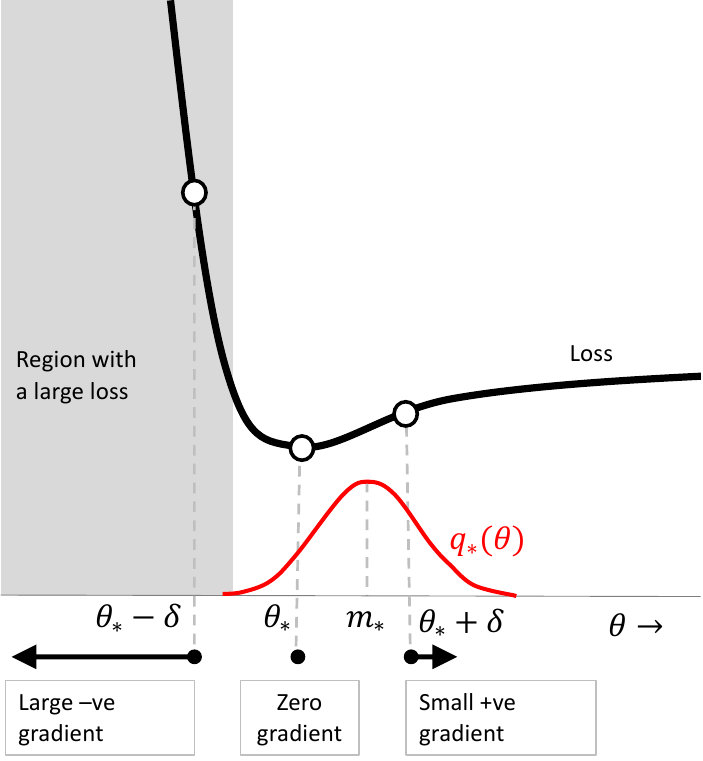}
        \label{fig:bayes_robust_1}} \hspace{1cm}
    \subfigure[]{\includegraphics[height=2.5in]{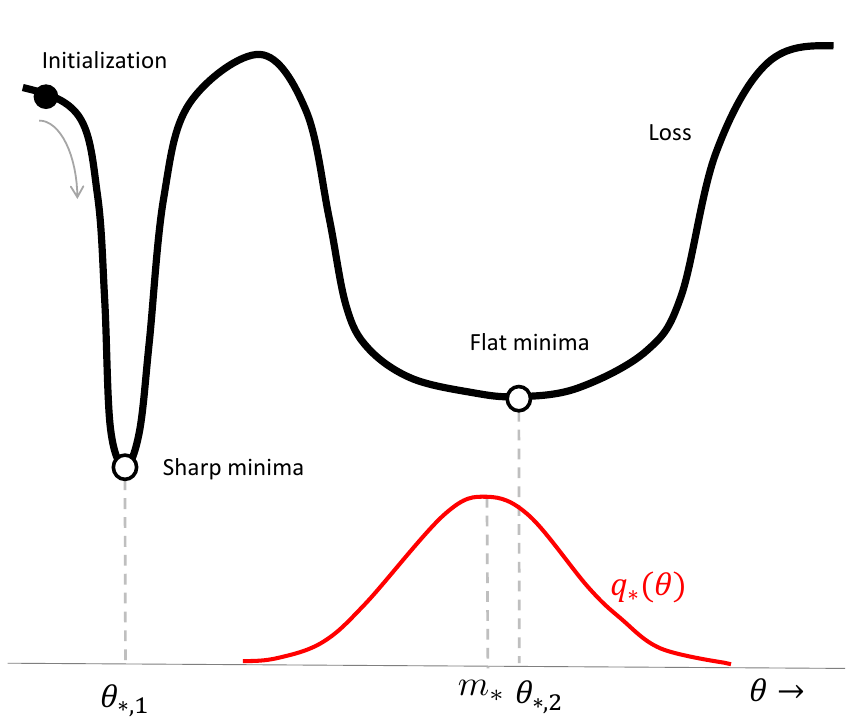}
        \label{fig:bayes_robust_2}}
    \caption{Bayesian solutions have similar robustness
            properties to flatter minima found in deep learning via
            stochastic algorithms. Panel (a): When the minimum lies
        right next to a `wall', the Bayesian solution shifts away from
        the wall (towards the flatter side) to avoid large losses
        under small perturbation. This is due to \updated{the averaging over
   $q_*(\vparam)$} in condition
        \cref{eq:conv_cond_1}. Panel (b): Given a sharp minimum vs a
        flat minimum, the Bayesian solution often prefers the flatter
        minimum, which is again due to \updated{the averaging over $q_*(\vparam)$}.}
    \label{fig:bayes_robust}
\end{figure}

OGN's uncertainty estimates can be improved further by 
relaxing the
    approximation made by the delta method and using Bayesian
    averaging through variational inference. This improves over
Laplace's method by using the stationarity conditions
(\cref{eq:conv_cond_1,eq:conv_cond_2}) where Bayesian averaging is
employed through an expectation over $q(\vparam)$ \citep{Opper:09}.
The variational solution is slightly different than Laplace's method
and is expected to be more robust due to the averaging over
$q_*(\vparam)$ in \cref{eq:conv_cond_1,eq:conv_cond_2}. Two such
situations are illustrated in \cref{fig:bayes_robust}, one
corresponding to an asymmetric loss where the Bayesian solution avoids
a region with extremely high loss (\cref{fig:bayes_robust_1}), and the
other one where it seeks a more stable solution involving a wide and
shallow minimum compared to a sharp and deep minimum
(\cref{fig:bayes_robust_2}). Such situations are hypothesised to exist
in deep-learning problems \citep{hochreiter1997flat,
    hochreiter1995simplifying, keskar2016large}, where a good
performance of stochastic optimization methods is attributed to their
ability to find shallow minima \citep{dziugaite2017computing}. Similar
strategies exist in stochastic search and global optimization
literature where a kernel is used to convolve/smooth the objective
function, like in Gaussian homotopy continuation
method~\citep{mobahi2015theoretical}, optimization by
smoothing~\citep{leordeanu2008smoothing}, graduated optimization
method \citep{hazan2016graduated}, and evolution strategies
\citep{huning1976evolutionsstrategie, wierstra2008natural}.
The stochastic noise in these methods is akin to the Bayesian
    noise injected through the distribution $q(\vparam)$. Unlike the
    `unregulated' noise in these methods, the benefit of the Bayesian
    approach is that the noise source $q(\vparam)$ can adapt itself
    from data by using the BLR.

The variational-Bayesian version is obtained by simply removing the
approximation due to the delta method from \cref{eq:don_minibatch},
going back to the original Newton variant of \cref{eq:von_1} (now with
minibatch Hessians),
\begin{align}
  \vparam_{t+1} \leftarrow \vparam_t -
  \rho_t \frac{1}{\vs_{t+1}} \circ \myexpect_{q_t}\sqr{ \widehat{\grad}_{\text{\vparam}}
  \barloss(\vparam) }, \quad\text{with} \quad
  \vs_{t+1} \leftarrow (1-\rho_t) \vs_t +
   \rho_t\,\, \myexpect_{q_t}\sqr{ \diag(\widehat{\grad}_{\text{\vparam}}^2 
  \barloss(\vparam) ) } ,
  \label{eq:von_minibatch}
\end{align}
where iterates $q_t(\vparam) = \gauss(\vparam|\vparam_t, \vS_t^{-1})$
are defined with precision matrix $\vS_t$ as a diagonal matrix with
$\vs_t$ as the diagonal. The iterates converge to an optimal diagonal
Gaussian candidate that optimizes \cref{eq:BayesP}.

Similarly to OGN, the update \cref{eq:von_minibatch} can be modified
to use the Gauss-Newton approximation \cref{eq:OGN_approx}, which is
easier to implement and can also exploit the deep-learning tricks to
find good solutions. The expectations can be implemented by a simple
`weight-perturbation',
\[ \myexpect_{q_t} [ \widehat{\grad}_{\vparam} \barloss(\vparam) ]
    \approx \widehat{\grad}_{\vparam} \barloss(\vparam_t +
    \vepsilon_t) \] where
$\vepsilon_t \sim \gauss(\vepsilon| 0, \diag(\vs_t))$ (multiple
samples can also be used). The resulting algorithm is referred to as
variational OGN or VOGN by \citet{khan2018fast}) and is shown to match
the performance of Adam and give better uncertainty estimates than
both Adam, OGN, and other Bayesian alternatives
\citep[Fig.~1]{osawa2019practical}. A version with the FIM of the
log-likelihood is proposed in \citet{dnn2gp} (algorithm is called
VOGGN). Variants that exploit Hessian structures using
Kronecker-factored \citep{zhang2018noisy} and low-rank
\citep{mishkin2018slang} are also considered in the literature. 

\subsection{Binary Neural Networks} 
\label{sec:binn}

Binary NNs have binary weights $\vparam \in \{-1,+1\}^P$, hence we
need to consider another class of the candidate distributions than the
Gaussian (we will choose Bernoulli). Binary NNs are natural candidates
for resource-constrained applications, like mobile phones, wearable
and IoT devices, but their training is challenging due to the binary
weights for which gradients do not exists. Our Bayesian scheme solves
this issue because gradients of the Bayesian objective are always
defined with respect to the parameters of a Bernoulli distribution,
turning a discrete problem into a continuous one. The discussion here
is based on the results of \cite{meng2020training} who used the BLR to
derive the BayesBiNN learning algorithm and connect it to the well
known Straight-Through-Estimator (STE) \citep{bengio2013estimating}.

The training objective of binary NN involves a discrete optimization problem
\begin{equation}
  \min_{\text{\vparam} \in \{-1,+1\}^P} \,\,
  \sum_{i=1}^N \loss( y_i, f_{\text{\vparam}}(\vx_i) ). 
   \label{eq:binn_obj}
\end{equation}
In theory, gradient based methods should not work for such problems
since the gradients with respect to discrete variables do not exist.
Still, the most popular method to train binary NN, the
Straight-Through-Estimator (STE) \citep{bengio2013estimating}, employs
continuous optimization methods and works remarkably well
\citep{courbariaux2015binaryconnect}. The method is justified based on
latent real-valued weights $\tvparam\in\real^P$ which are discretized
at every iteration to get binary weights $\vparam$, while the
gradients used to update the latent weights $\tvparam$, are computed
at the binary weights $\vparam$, as shown below,
\begin{equation}
  \tvparam_{t+1} \leftarrow  \tvparam_t -
  \alpha \sum_{i\in\minibatch_t} \grad_{\text{\vparam}}
  \loss(y_i, f_{\text{\vparam}_t}(\vx_i)),
  \text{ where } \vparam_t \leftarrow \text{sign}(\tvparam_t)
  \label{eq:STE}
\end{equation}
It is not clear why the gradients computed at the binary weights help
the search for the minimum of the discrete problem, and many studies
have attempted to answer this question, e.g., see
\cite{yin2019understanding, alizadeh2018empirical}. The Bayesian
objective in \cref{eq:BayesP} rectify this issue since the expectation
of a discrete objective is a continuous objective with respect to the
parameters of the distributions, justifying the use of gradient based
methods.

We will now show that using Bernoulli candidate distributions in the
BLR gives rise to an algorithm similar to \cref{eq:STE}, justifying the
application of the STE algorithm to solve a discrete optimization
problem. Specifically, we choose $q(\param_j=1)=p_j$ where
$p_j\in[0,1]$ is the success probability. We also assume a mean-field
distribution to get $q(\vparam) =\prod_{j=1}^{P} q(\param_j)$. Since
the Bernoulli distribution is a minimal exponential-family with a
constant $h(\vparam)$, we use the BLR in \cref{eq:bayes_learn_rule}
with $\natparam_j = 1/2 \log({p_{j}}/(1-p_{j}))$ and
$\mu_{j} = 2p_j-1$, to get
\begin{align}
  \vnatparam_{t+1} &\leftarrow (1- \rho_t) \vnatparam_t -
                     \rho_t \grad_{\vmeanparam}
                     \myexpect_{q_t}
                     \sqr{ \sum_{i\in\minibatch_t}
                     \loss( y_i, f_{\text{\vparam}}(\vx_i) )}.
   \label{eq:bayes_learn_binn}
\end{align}
The challenge now is to compute the gradient with respect to
$\vmeanparam$, but it turns out that approximating it with the
Concrete distribution \citep{maddison2016concrete,
    jang2016categorical} recovers the STE step.

Concrete distribution provides a way to relax a discrete random
variable into a continuous one such that it becomes a deterministic
function of a continuous random variable. Following
\citet{maddison2016concrete}, the continuous random variable, denoted
by $\hat{\param}^{(\tau)} \in (-1, 1)$, are constructed from the
binary variable $\param \in \{-1, 1\}$ by using iid uniform random
variables $\epsilon \in (0,1)$ and using the function $\tanh(\cdot)$,
\begin{equation}
   \hat{\param}^{(\tau)} = \tanh\rnd{\frac{ \lambda +
       \delta(\epsilon)}{\tau} } ,
   \text{ where } \delta(\epsilon) = \half \log\rnd{\frac{\epsilon}{1-\epsilon}},
   \label{eq:param_b_tau}
\end{equation}
with $\tau>0$ as the temperature parameter. As $\tau \to 0$, the
function approaches the sign function used in STE step
(\cref{eq:STE}):
$\hat{\param}^{(0)} = \text{sign}( \lambda + \delta(\epsilon))$.

The gradient too can be approximated by the gradients with respect to
the relaxed variables
\begin{align}
  \grad_{\text{\vmeanparam}} \myexpect_{q} \sqr{ \loss( y_i,
  f_{\text{\vparam}}(\vx_i) )}
  \approx
  \grad_{\text{\vmeanparam}}
  \loss( y_i, f_{\hat{\text{\vparam}}^{(\tau)}}(\vx_i) ) =
   \vs \circ \sqr{ \grad_{\hat{\text{\vparam}}^{(\tau)}}
   \loss( y_i, f_{\hat{\text{\vparam}}^{(\tau)}}(\vx_i) ) },  
\end{align}
where we used chain rule in the last step and $\vs$ is a vector of
\[ s_j = \grad_{\meanparam_j} \param_j^{(\tau)} = \frac{1}{\tau} \sqr{ \frac{
   1-(\hat{\param}_j^{(\tau)})^2}{ 1- \tanh^2(\natparam_j)} }.\]
By renaming $\vnatparam$ by $\tvparam$ and $\hat{\vparam}^{(\tau)}$ by
$\vparam$, the rule in \cref{eq:bayes_learn_binn} becomes
\begin{align}
  \tvparam_{t+1} \leftarrow (1- \rho_t) \tvparam_t - \rho_t
   \vs_t \circ \sqr{ \sum_{i\in\minibatch_t} \grad_{\text{\vparam}}
   \loss( y_i, f_{\text{\vparam}_t}(\vx_i) ) }, \text{ where }
  &
    \vparam_t \leftarrow \tanh\rnd{\frac{1}{\tau} \rnd{
    \tvparam_t + \vdelta(\vepsilon_t)} }
    \label{eq:bayes_learn_binn_final}
\end{align}
with
$\vs_t \leftarrow (1 - \vparam_t^2)/(\tau (1-\tanh^2(\tvparam_t)))$,
and $\vepsilon_{t}$ is a vector of $P$ iid samples from a uniform
distribution. \citet{meng2020training} refer to this algorithm as
BayesBiNN.

BayesBiNN is similar to STE shown in \cref{eq:STE}, but computes
natural-gradients (instead of gradients) at the relaxed parameters
$\vparam_t$ which are obtained by adding noise $\vdelta(\vepsilon_t)$
to the real-valued weights $\tvparam_t$. The gradients are
approximations of the expected loss and are well defined. By setting
noise to zero and letting the temperature $\tau \to 0$, the
$\tanh(\cdot/\tau)$ becomes the sign-function and we recover the
gradients used in STE. This limit is when the randomness is ignored
and is similar to the delta method used in earlier sections. The
random noise $\vdelta(\vepsilon_t)$ and the non-zero temperature
$\tau$ enables a softening of the binary weights, allowing for
meaningful gradients. BayesBiNN also provides meaningful latent
$\tvparam$ which are now the natural parameters of the Bernoulli
distribution.

Unlike STE, the updates employs a exponential smoothing which is a
direct consequence of using the entropy term in the Bayesian
objective. With discrete weights, optimum of \cref{eq:binn_obj} could
be completely unrelated to an STE solution $\vparam_*$, for example,
the one that yields zero gradients
$\sum_i \grad_{\text{\vparam}} \loss(y_i, f_{\text{\vparam}_*}(\vx_i))
= 0$. In contrast, the BayesBiNN solution minimizes the well-defined
Bayes objective of \cref{eq:BayesP}, whose optimum is characterized by
the optimality condition \cref{eq:fixed_pt} directly relating the
optimal $\tvparam_*$ to the relaxed variables $\vparam_*$:
\begin{align}
   \frac{\tvparam_{*}}{\vs_*} \approx \sum_{i=1}^N
  \grad_{\text{\vparam}} \loss(y_i, f_{\text{\vparam}_{*}}(\vx_i)) .
\end{align}
The BayesBiNN solution $\vparam_{*}$ is in general different from the
one from the STE algorithm, but when temperature $\tau \to 0$ we
expect the two solutions to be similar whenever $\vs_* \to \infty$ (as
the left-hand-side in the equation above goes to zero). In general, we
expect the BayesBiNN solution to have robustness properties similar to
the ones discussed for \cref{fig:bayes_robust}. The exponential
smoothing used in BayesBiNN is similar to the update of
\citet{Helwegen2019} but their formulation lacks a well-defined
objective.



\section{Probabilistic Inference Algorithms from the Bayesian Learning Rule}
\label{sec:approx_bayes}

Algorithms for inference in probabilistic graphical models can be
derived from the BLR, by setting the loss function to be the log-joint
density of the data and unknowns. Unknowns could include both latent
variables $\vlat$ and model parameters $\vparam$. The class
${\mathcal Q}$ is chosen based on the form of the likelihood and
prior. We will first discuss the conjugate case, where BLR
reduces to Bayes' rule, and then discuss expectation maximization and
variational inference (VI) where structural approximations to the
posterior are employed.

\subsection{Conjugate Models and Bayes' Rule}
\label{sec:conj_bayes}

We start with one of the most popular category of Bayesian models
called exponential-family conjugate models. We consider $N$ i.i.d. data
$\vy_i$ with likelihoods $p(\vy_i|\vparam)$ (no latent
variables) and prior $p(\vparam)$. \updated{Conjugacy implies the existence of
sufficient statistics $\vT(\vtheta)$ such that both the likelihood and prior take an exponential form shown below}
\begin{equation}
   p(\vy_i|\vparam) \propto e^{\myang{\tvlambda_i(\vy_i),
    \vT(\vparam)}},
  \quad \text{ and }
   \quad p(\vparam) \propto e^{\myang{\vnatparam_0, \vT(\vparam)} },
  \nonumber
\end{equation}
for some $\tvlambda_i(\vy_i)$ (a function of $\vy_i$), and
$\vlambda_0$.
\updated{For example, for ridge regression discussed in \cref{sec:ridge}, the likelihoods and prior are 
\begin{align*}
   p(y_i|\vparam) &= \gauss(y_i|\vx_i^\top \vparam, 1) \propto  e^{(y_i\vx_i)^\top \vparam + \trace\rnd{- \half \vx_i\vx_i^\top \rnd{\vparam \vparam^\top}} - \half y_i^2},\\
   p(\vparam) &= \gauss(\vparam|0, \vI/\delta) \propto e^{\trace\rnd{- \half\delta\vI \rnd{ \vparam\vparam^\top} }},
\end{align*}
respectively, and, because $\vT(\vparam) = (\vparam, \vparam\vparam^\top)$, we have the following pairs
\begin{equation*}
   \begin{array}{ll}
      \begin{array}{ll}
         \tvlambda_i^{(1)}(y_i) &= y_i\vx_i, \\ 
         \tvlambda_i^{(2)}(y_i) &= -\half \vx_i\vx_i^\top,
       \end{array}
       &\quad
       \begin{array}{ll} 
         \vnatparam_0^{(1)} &= 0, \\
         \vnatparam_0^{(2)} &=-\half \delta\vI.
       \end{array}
   \end{array}
\end{equation*}}
The quantity $\tvlambda_i(\vy_i)$ is often interpreted as the
sufficient statistics of $\vy_i$, but it can also be seen as the
natural parameter of the likelihood with respect to
$\vT(\vparam)$.~The posterior is then obtained by a simple addition of these natural parameters, as shown below,
\begin{displaymath}
   p(\vparam|\vy) \,\,\propto  p(\vparam) \prod_i p(\vy_i|\vparam)
   \,\,\propto \exp\rnd{ \myang{\vnatparam_0 + \sum_i
        \tvlambda_i(\vy_i), \vT(\vparam)}},
\end{displaymath}
\updated{where $\vy = (\vy_1,\vy_2,\ldots,\vy_N)$. 

The natural-parameter of the posterior can be obtained from the BLR. To see this, we first note that the loss needed in the BLR is the negative of the log-joint density 
\[
   \barloss(\vparam) = - \log p(\vy, \vparam) =  - \log p(\vparam) + \sum_{i=1}^N \sqr{ -\log p(\vy_i|\vparam)}.
\]
The natural parameter of the posterior is recovered from the fixed point of the BLR (\cref{eq:fixed_pt_specific}),
\begin{equation}
   \begin{split}
      \vnatparam_* 
      = \grad_{\text{\vmeanparam}}\myexpect_{q_*}[\log p(\vparam)] + \sum_i \grad_{\text{\vmeanparam}}\myexpect_{q_*}[\log p(\vy_i|\vparam)] 
      = \vnatparam_0 + \sum_i \tvlambda_i(\vy_i), 
   \end{split}
   \label{eq:conj_blr_update}
\end{equation}
which is equivalent to running the BLR for just one-iteration with $\rho_t = 1$. For ridge regression, for instance, the above equation results in the natural gradients in \cref{eq:natgrad_ridge} from which the ridge-regression solution can be recovered, as shown in \cref{sec:ridge}. 
}

%

\updated{In general, for such conjugate exponential-family models, the BLR
yields the exact posterior by a direct application of \cref{eq:fixed_pt_specific}.
Applying Bayes' rule there is equivalent to a single
step of \cref{eq:bayes_learn_rule} with $\rho_t = 1$.}~This includes popular Gaussian linear models~\citep{roweis1999unifying}, such as Kalman
filter/smoother, probabilistic Principle Components Analysis (PCA),
factor analysis, mixture of Gaussians, and hidden Markov models, as
well as nonparameteric models, such as Gaussian process (GP)
regression \citep{Rasmussen06}. 
For all such cases, the BLR gives yields the natural parameter of the posterior in just one step. 
\updated{From a computational point of view, it does not offer any advice to obtain
marginal properties in general, and, similarly to Bayes' rule, efficient techniques, such as
message passing, may be needed to perform the necessary computation.}

\subsection{Expectation Maximization}
\label{sec:em}

We will now derive Expectation Maximization (EM) from the BLR
with coordinate-wise updates with $\rho_t=1$, similarly to
    the conjugate case. In the subsequent sections, this connection is
    used to derive new stochastic VI algorithms where sometimes $\rho_t < 1$ is
    used.

EM is a popular algorithm for parameter estimation with latent
variables~\citep{dempster1977maximum}. Given the joint distribution
$p(\vy, \vlat| \vparam)$ with latent variables $\vlat$ and parameters
$\vparam$, the EM algorithm computes the posterior
$p(\vlat|\vy,\vparam_*)$ as well as parameter $\vparam_*$ that
maximizes the marginal likelihood $p(\vy|\vparam)$. The algorithm can
be seen as coordinate-wise BLR updates with $\rho_t = 1$ to update
candidate distribution that factorizes across $\vlat$ and $\vparam$,
\begin{equation}
  q_t(\vlat,\vparam) = q_t(\vlat) q_t(\vparam)  \,\,\propto \,\,
  e^{  \myang{\vnatparam_t, \vT_\lat(\vlat)}} \gauss(\vparam|\vparam_t, \vI),
\end{equation}
where $\vparam_t$ denotes the mean of the Gaussian. The loss function
is set to $\barloss(\vlat, \vparam) = -\log p(\vy,\vlat|\vparam)$.

To ease the presentation, assume the likelihood, prior, and joint 
are expressed as
\begin{align}
   p(\vy|\vlat, \vparam) &\propto \exp\rnd{
    \myang{\tvlambda_1(\vy,\vparam), \vT_\lat(\vlat)}}, 
 &p(\vlat|\vparam) \propto \exp\rnd{
    \myang{\vnatparam_0(\vparam), \vT_\lat(\vlat)} }, \nonumber\\
 p(\vy,\vlat| \vparam) &\propto \exp\rnd{ 
    \myang{\vT_{y\lat}(\vy,\vlat), \vparam} -
    A_{y\lat}(\vparam) }, &
    \label{eq:em_model}
\end{align}
for some functions $\tvlambda_1(\cdot,\cdot)$ and
$\vnatparam_0(\cdot)$ (similarly to the previous section), and $\vparam$
is the natural parameter of the joint. The likelihood
and prior are expressed in terms of sufficient statistics
$\vT_\lat(\vlat)$ while the joint is with $\vT_{y\lat}(\cdot,\cdot)$
\citep{winn2005variational}. The EM algorithm then takes a simple form
where we iterate between updating $\vnatparam$ and $\vparam$
\citep{banerjee2005clustering},
\begin{displaymath}
    \text{E-step:}\quad \vnatparam_{t+1} \leftarrow
    \tvlambda_1(\vy,\vparam_t) + \vnatparam_0(\vparam_t), 
    \qquad 
    \text{M-step:}\quad \vparam_{t+1} \leftarrow \nabla A^*_{y\lat}
    \rnd{ \myexpect_{q_{t+1}} [ \vT_{y\lat}(\vy, \vlat) ] }. 
\end{displaymath}
Here, $A_{y\lat}^*(\cdot)$ is the Legendre transform, and
$q_{t+1}(\vlat) = p(\text{\vlat}|\vy,\vparam_{t+1})$ is the outcome of E-step.

The two steps are obtained from the BLR by using the delta method to
approximate the expectation with respect to $q_t(\vparam)$. For the
E-step, we assume $q_t(\vparam)$ fixed, and use
\cref{eq:fixed_pt_specific,eq:conj_blr_update} and the delta method,
\begin{displaymath}
   \vnatparam_{t+1} \leftarrow \left. \grad_\vmeanparam 
      \myexpect_{q_t(\vlat)q_t(\vparam)} [ -\barloss(\vlat,\vparam) ] 
   \right\vert_{\vmeanparam = \vmeanparam_t} =
   \myexpect_{q_t(\vparam)} [
   \tvlambda_1(\vy,\vparam) + \vnatparam_0(\vparam) ]
   \approx \tvlambda_1(\vy,\vparam_t) + \vnatparam_0(\vparam_t) 
\end{displaymath}
For the M-step, we use the stationarity condition similar to
\cref{eq:conv_cond_1} but now with the delta method with respect to
$q(\vparam)$ (note that $\vparam_{t+1}$ is both the expectation and
natural parameter),
\begin{displaymath}
    \left. \grad_\vparam \myexpect_{q_{t+1}(\vlat)}
      \sqr{ \barloss(\vlat,\vparam)  }\right\vert_{\vparam =
        \vparam_{t+1}} = 0
\end{displaymath}
A simple calculation using the joint $p(\vy,\vlat)$ will show that
this reduces to the M-step.

The derivation is easily extended to the generalized EM iterations.
The coordinate-wise strategy also need not be employed and the BLR
steps can also be run in parallel. Such scheme resembles the
generalized EM algorithm and usually converges faster. Online schemes
can be obtained by using stochastic gradients, similar to those
considered in \citet{titterington1984recursive, neal1998view,
    sato1999, cappe2009line, delyon1999convergence}. More recently,
\citet{pmlr-v124-amid20a} use a divergence-based approach and
\citet{pmlr-v130-kunstner21a} prove attractive convergence properties
using a mirror-descent formulation similar to ours. Next, we derive a
generalization of such online schemes, called the stochastic
variational inference.

\subsection{Stochastic Variational Inference and Variational Message
    Passing}
\label{sec:svi}

Stochastic variational inference (SVI) is a generalization of EM,
first considered by \citet{sato2001online} for the conjugate case and
extended by \citet{hoffman2013stochastic} to conditionally-conjugate
models (a similar strategy was also proposed by \citet{Honkela:11}).
We consider the conjugate case due to its simplicity. The model is
assumed to be similar to the EM case as in \cref{eq:em_model}, but with
$N$ iid data examples $\vy_i$ associated with one latent vector
$\vz_i$ each and a conjugate prior $p(\vparam|\valpha)$ with
$\valpha = (\valpha_1, \alpha_2)$ as prior parameters,
\begin{align}
  p(\vy_i|\vlat_i, \vparam) &\propto \exp\rnd{
                              \myang{\tvlambda_i(\vy_i,\vparam), \vT_i(\vlat_i)}}, 
  && p(\vlat_i|\vparam) \propto \exp\rnd{ \myang{\vnatparam_0(\vparam),
     \vT_i(\vlat_i)} }, \nonumber\\
  p(\vy_i, \vlat_i| \vparam) &\propto \exp\rnd{ \myang{\vT_{y\lat}(\vy_i,\vlat_i), \vparam} - A_{y\lat}(\vparam) },
  && p(\vtheta|\valpha) = h_0(\vtheta) \exp \rnd{ \sqr{\myang{\valpha_1, \vparam}
     - \alpha_2 A_{y\lat}(\vparam) }}.
     \label{eq:svi_prior}
\end{align}
Due to conjugacy, each conditional update can be done coordinate wise.
A common strategy is to assume a mean-field approximation
\[
  q(\vlat_1,\ldots, \vlat_N, \vparam) = q(\vparam) \prod_i q(\vlat_i)
   \,\,\propto \,\, e^{ \vnatparam_0^{(1)} - \natparam_0^{(2)}
    A_{y\lat}(\vparam) } \prod_i e^{ \myang{\vnatparam_i,
      \vT_i(\vlat_i)}},
\]
where $\vnatparam_i$ is the natural parameter of $q(\vlat_i)$, and
$(\vnatparam_0^{(1)}, \natparam_0^{(2)})$ is the same for $q(\vparam)$.

Coordinate-wise updates when applied to general graphical model are
known as variational message passing \citep{winn2005variational}. These
updates are a special case of BLR, applied coordinate wise with
$\rho_t=1$. The derivation is almost identical to the one used for the
EM case earlier, therefore omitted. One important difference is that,
we do not employ the delta method for $q(\vparam)$ and explicitly
carry out the marginalization which has a closed-form expression due
to conjugacy. The update for $q(\vparam)$ is therefore replaced by its
corresponding conjugate update.

Stochastic variational inference employs a special update where, after
every $q(\vlat_i)$ update with $\rho_t = 1$, we update $q(\vparam)$
but with $\rho_t<1$. Clearly, this is also covered as a special case
of the BLR. The BLR update is more general than these strategies since
it applies to a wider class of non-conjugate models, as discussed
next.

\subsection{Non-Conjugate Variational Inference}
\label{sec:ncvi}

Consider a graphical model
$p(\vx) \propto \prod_{i\in {\mathcal I}} f_i(\vx_i)$ where $\vx$
denotes the set of latent variables $\vz$ and parameters $\vparam$ and
$f_i(\vx_i)$ is the $i$'th factor operating on the subset $\vx_i$
indexed by the set ${\mathcal I}$. The BLR can be applied to estimate
a minimal exponential-family (assuming a constant $h(\vparam)$),
\begin{equation}
    \vnatparam_{t+1} \leftarrow (1- \rho_t) \vnatparam_t +
    \rho_t \sum_{i\in{\mathcal I}} \tvlambda_i(\vmeanparam_t),
    \text{ where }
    \tvlambda_i(\vmeanparam_t) = \left. \grad_{\boldsymbol{\meanparam}}
      \myexpect_q \sqr{\log f_i(\vx_i)} \right\vert_{\vmeanparam = \vmeanparam_t},
      \label{eq:blr_global}
\end{equation}
\updated{and} $\vmeanparam_t = \grad_\vnatparam A(\vnatparam_t)$. We
    already encountered the quantities $\tvlambda_i(\vmeanparam_t)$
    for conjugate models in \cref{sec:conj_bayes} where they are
    referred to as the natural parameter of the factors. They can also
    be interpreted as ``local'' messages that are aggregated to obtain
    the ``global'' natural parameter $\vnatparam_{t+1}$.

The update is quite flexible and can also be applied to
    other graphical models, such as Bayesian networks, Markov random
    fields, and Boltzmann machines. We can also view it as an update
over distributions, by multiplying by $\vT(\vx)$ followed by
exponentiation,
\begin{equation}
    q_{t+1}(\vx) \propto q_t(\vx)^{(1- \rho_t)}
    \sqr{ \prod_{i\in {\mathcal I}} \exp{ \rnd{ \myang{
                \tvlambda_i(\vmeanparam_t) ,
                \vT(\vx)}}} }^{\rho_t}
    \label{eq:blr_dist}
\end{equation}
and at convergence
\begin{displaymath}
    q_*(\vx) \propto \prod_{i \in {\mathcal I}} \exp{ \rnd{\myang{
            \tvlambda_i(\vmeanparam_*) ,
            \vT(\vx)}}},
\end{displaymath}
where $q_*(\vx)$ is the optimal distribution with natural and
expectation parameters as $\vnatparam_*$ and $\vmeanparam_*$. This
update has three important features.
\begin{enumerate}
\item The optimal $q_*(\vx)$ has the same structure as $p(\vx)$ and
    each term in $q_*(\vx)$ yields an approximation to those in
    $p(\vx)$,
    \begin{displaymath}
        \log f_i(\vx_i) \approx \myang{ 
            \tvlambda_i(\vmeanparam_*) , \vT(\vx)} + c_i
    \end{displaymath}
    for some constant $c_i$. The quantity $\tvlambda_i(\cdot)$ can be
    interpreted as the natural parameter of the local approximation.
    Such local parameters are often referred to as the site parameters
    in expectation propagation (EP) where their computations are often
    rather involved. In our formulation they are simply the natural
    gradients of the expected log joint density. See \citet[Sec.~3.4
    and 3.5]{chang2020fast} for an example comparing the two.
    The above local approximations are inherently
        superior to the local-variational bounds, for example,
        quadratic bounds used in logistic regression
        \citep{Jaakkola96b, Khan10, khan2012stick, emtThesis}, and
        also to the augmentation strategies
        \citep{girolami2006variational, pmlr-v39-klami14}. 
        Gaussian posteriors give rise to quadratic surrogates to the loss function. 
        These can be seen as alternatives to those obtained using Taylor's expansion.
        The advantage of our surrogates is that they are
        not tight locally, rather globally \citet{Opper:09}.

\item The update in \cref{eq:blr_dist} yields both the local
    $\tvlambda_i(\cdot)$ and global $\vnatparam$ parameters
    simultaneously, which is unlike the message-passing strategies
    that only compute the local messages. In fact, \cref{eq:blr_dist}
    can also be reformulated entirely in terms of local parameters
      (denoted by $\tvlambda_{i}^{(t+1)}$ below with a slight abuse of notation),
    \begin{equation}
            \vnatparam_{t+1} \leftarrow \sum_{i\in{\mathcal I}}
            \tvlambda_{i}^{(t+1)}, \text{ where } \tvlambda_i^{(t+1)}
            \leftarrow (1- \rho_t) \tvlambda_i^{(t)} + \rho_t
            \tvlambda_i(\vmeanparam_t),
            \label{eq:blr_local}
    \end{equation}
    This formulation was discussed by \citet{khan2017conjugate} who
    show this to be a generalization of the non-conjugate variational
    message passing algorithm \citep{knowles2011non}. The form is
    suitable for a distributed implementation using a message-passing
    framework.

 \item The update applies to both conjugate and non-conjugate factors, \updated{but also for both} probabilistic and deterministic variables. Automatic
    differentiation tools can be easily integrated. \updated{Due to such properties, the BLR offers an attractive framework} for practical probabilistic programming
    \citep{van2018introduction}.
\end{enumerate}
We have argued that natural gradients play an important
role for probabilistic inference, both for conjugate models or within
those using message-passing frameworks. Still, their importance is
somehow missed in the community. Natural gradients were first
introduced in this context by \citet{khan2017conjugate, khan2018fast},
although \citet{salimans2013fixed} also mention a similar condition
without an explicit reference to natural gradients.
\citet{sheth2016monte, sheth2019algorithms} discuss an update similar
to the BLR for a specific case of two-level graphical models.
\citet{salimbeni2018natural} propose natural-gradient strategies for
sparse GPs by using automatic differentiation which is often slower
(the original work by \cite{10.5555/3023638.3023667}, is \updated{also} a
special case of the BLR). A recent generalization to structured
natural-gradient by \citet{lin_stucturedngd} is also worth noting.

We will end the section by discussing the connections of the BLR to
the algorithms used in online learning, where the loss functions are
observed sequentially at every iteration $t$,
\begin{displaymath}
   q_{t+1}(\vparam) = \argmin_{q(\text{\vparam})} \quad
   \myexpect_{q} \sqr{  
    \loss(y_t, f_{\text{\vparam}}(\vx_t)) }
    + \frac{1}{\rho_t} \dkls{}{q(\vparam)}{q_t(\vparam)}.
\end{displaymath}
The resulting updates are called Exponential-Weight (EW) updates
\citep{littlestone1994weighted, Vovk:1990:AS:92571.92672}, which are
closely related to Bayes' rule. \citet{hoeven2018many} show that, by
approximating the first term by a surrogate (linear/quadratic) at the
posterior mean (the delta method), many existing online learning
algorithms can be obtained as special cases. This includes, online
gradient descent, online Newton step, online mirror descent, among
many others. This derivation is similar to the BLR which, when applied
here, is slightly more general, not only due to the use of expectated
loss, but also because the surrogate
choice is automatically obtained by the posterior approximations. 

An advantage of this connection is that the theoretical guarantees derived
in online learning can be translated to the BLR, and consequently to
all the learning-algorithms derived from it.
We will also note that the local and global BLR updates presented in 
\cref{eq:blr_global,eq:blr_local} are similar to the `greedy' and `lazy' updates used in online learning \citep{hoeven2018many}. The conjugate
case discussed in earlier sections are very similar to the algorithms
proposed for online learning by \citet{azoury2001relative}, which are
concurrent to a similar proposal in the Bayesian community by
\citet{sato2001online}.



\section{Discussion}
\label{sec:discussion}

To learn, we need to extract useful information from new data and
revise our beliefs. To do this well, we must reject false information
and, at the same time, not ignore any (real) information. A good
learning algorithm too must possess these properties. We expect the
same to be true for the algorithms that have ``stood the test of
time'' in terms of good performance, even if they are derived through
empirical experimentation and intuitions. If there exist an optimal
learning-algorithm, as it is often hypothesized, then we
might be able to view these empirically good learning-algorithms as
derived from this common origin. They might not be perfect but we
might expect them to be a reasonable approximation of the optimal
learning-algorithm. All in all, it is possible that all such
algorithms optimize similar objectives and use similar strategies on
how the optimization proceeds.

In this paper we argue for two ideas. The first is that the
learning-objective is a Bayesian one and uses the variational
formulation by \cite{art669} in \cref{eq:BayesP}. The objective tells
us \emph{how} to balance new information with the old one, resulting
ultimately in Bayes theorem as the optimal information processing
rule. The power of the variational formulation, is that it also shows
how to process the information when we have limited abilities, for example, 
in terms of computation, to extract the relevant information. The role
of $q(\vtheta)$ is to define how to represent the knowledge, for which
exponential families and mixtures thereof, are natural choices that
have shown to balance complexity while still being practical.

The second idea is the role natural gradients play in the learning
algorithms. Bayes' rule in its original form has no opinion about
them, but they are inherently present in all solutions of the Bayesian
objective. They give us information about the learning-loss landscape,
sometimes in the form of the derivatives of the loss and sometimes as
the messages in a graphical model. Our work shows that all these
seemingly different ideas, in fact, have deep roots in information
geometry that is being exploited by natural gradients and then in
Bayesian Learning Rule (BLR) we proposed.

The two ideas go together as a symbiosis into the BLR. By a long
series of examples from optimisation, deep learning, and graphical
models, we demonstrate that classical algorithms such as ridge
regression, Newton’s method, and Kalman filter, as well as modern
deep-learning algorithms such as stochastic-gradient descent, RMSprop,
and Dropout, all can be derived from the proposed BLR. The BLR
gives a unified framework and understanding of what various (good)
learning algorithms do, how they differ in approximating the target
and then also how they can be improved. The main advantage of the BLR,
is that we now have a principled framework to approach new learning
problems, both in terms of what to aim for and also how to get there.



\section*{Acknowledgement}
M.~E.~Khan would like to thank many current and past colleagues at
RIKEN-AIP, including W.~Lin, D.~Nielsen, X.~Meng, T.~M{\"o}llenhoff
and P.~Alquier, for many insightful discussions that helped shape
parts of this paper. \updated{We also thank the anonymous reviewers for their feedback to improve the presentation.}

\appendix

\section{Bayesian Inference as Optimization}
\label{app:bayes_as_opt}

Bayesian inference is a special case of the Bayesian learning problem.
Although this follows directly from the variational formulation by
\cite{art669}, it might be useful to provide some more motivation.
Bayesian inference corresponds to a log-likelihood loss
$\loss(y,f_{\mathbf{\param}}(\vx)) = -\log
p(y|f_{\mathbf{\param}}(\vx))$. With conditional independent data and
prior $p(\vparam)$, the posterior distribution is
$p(\vparam|\data) = p(\vparam)/\mathcal{Z}(\data) \prod_{i=1}^N
p(\vy_i|f_\param(\vx_i))$. The minimizer of the Bayesian learning
problem recovers this posterior distribution. This follows directly
from reorganizing the Bayesian objective \cref{eq:BayesP}:
\begin{align}
  \mathcal{L}(q) &= - \myexpect_{q(\mathbf{\param})}
                   \sqr{ \sum_{i=1}^N \log
                   p(y_i|f_{\mathbf{\param}}(\vx_i)) }
                   + \dkls{}{q(\vparam)}{p(\vparam)}\\
                 &= \myexpect_{q(\mathbf{\param})}
                   \sqr{ \log \frac{q(\vparam)}{
                   \frac{p(\vparam)}{\mathcal{Z}(\mathcal{D})}
                   \prod_{i=1}^N p(y_i|f_{\mathbf{\param}}(\vx_i))} }
                   +
                   \log \mathcal{Z}(\data)\\
                 &= \dkls{}{q(\vparam)}{p(\vparam|\data)} + \log \mathcal{Z}(\data)
\end{align}
Choosing $q(\vparam)$ as $p(\vparam|\data)$ minimizes the above
equation since $\mathcal{Z}(\data)$ is a constant, and the KL
divergence is zero.

\section{Natural Gradient of the Entropy}
\label{app:natgrad_entropy}

Here, we show that $\natgrad_\vnatparam \entropy(q) = - \vnatparam$
when $h(\vparam)$ is a constant. First, we write
%
\begin{equation}
    \begin{split}
       - \natgrad_\vnatparam \entropy(q) &= \grad_{\vmeanparam} \myexpect_q[ \log q(\vparam) ] \\
       &= \nabla_{\vmeanparam} \sqr{ \myang{ \vnatparam, \vmeanparam
            } - A(\vnatparam) } +
        \nabla_{\vmeanparam} \myexpect_q[ \log h(\vparam)] \\
        &= \vnatparam + \rnd{ \grad_{\vnatparam}^2
            A(\vnatparam)}^{-1}
        \sqr{ \vmeanparam - \nabla_{\vnatparam} A(\vnatparam)} +
        \nabla_{\vmeanparam} \myexpect_q[ \log h(\vparam)] \\
        &= \vnatparam + \nabla_{\vmeanparam} \myexpect_q[ \log
        h(\vparam)],
    \end{split}
\end{equation}
where in the second line we use the definition of $q(\vparam)$ and in the third line we use chain-rule and the fact that $\vmu = \nabla_{\vnatparam} A(\vnatparam)$.
For constant $h(\vparam)$, this reduces to $\vnatparam$, giving us the result.


\section{The Delta Method}
\label{app:delta}

In the delta method \citep{dorfman1938note, ver2012invented}, we approximate the expectation of a
function of a random variable by the expectation of the function's 
Taylor expansion. For example, given
a distribution $q(\vparam)$ with mean $\vm$, we can use the
first-order Taylor approximation to get,
\begin{displaymath}
    \myexpect_q[ f(\vparam) ] \approx \myexpect_q \sqr{ f(\vm) + (\vparam
        - \vm)^\top \left. \nabla_{\vparam} f(\vparam) \right\vert_{\vtheta = \vm} } \,\, \approx f(\vm)
\end{displaymath}
We will often use this approximation to approximate the expectation of gradient/Hessian at their values at a single point (usually the mean).
For example, we can use a zeroth-order expansion to get
\begin{displaymath}
   \myexpect_q[ \nabla_\vparam f(\vparam) ] \approx  \left.
      \nabla_\vparam f(\vparam) \right\vert_{\vparam=\vm}
   \qquad
   \myexpect_q[ \nabla^2_\vparam f(\vparam) ] \approx \left. \nabla_\vparam^2 f(\vparam) \right\vert_{\vparam=\vm}.
\end{displaymath}

A better derivation (to get the same result) would be use higher-order expansion. For example, by using the first-order Taylor expansion, we approximate below the expectation of the gradient by the gradient at the mean,
\begin{align}
   \myexpect_q[ \nabla_\vparam f(\vparam) ] &= \nabla_\vm \myexpect_q[ f(\vparam) ]   \nonumber\\
    &\approx \nabla_\vm \myexpect_q \sqr{ f(\vm) + (\vparam
        - \vm)^\top \left. \nabla_{\vparam} f(\vparam) \right\vert_{\vtheta = \vm} } \nonumber\\
    &= \left.
      \nabla_\vparam f(\vparam) \right\vert_{\vparam=\vm},  
      \label{eq:firstderiv}
\end{align}
where the first line is due to Bonnet's theorem (\cref{app:bp}) and the second line is simply the first-order Taylor expansion.
We refer to this as the first-order delta method.
Similar applications have been used in variational inference \citep{blei2007correlated, braun2010variational, WangBlei}.

Similarly, by using the second-order Taylor expansion, we approximate below the expectation of the Hessian by the Hessian at the mean,
\begin{align}
   \myexpect_q[ \nabla_\vparam^2 f(\vparam) ] &= 
    2 \nabla_{\vS^{-1}} \myexpect_q[ f(\vparam) ]  \nonumber\\
   &\approx 2 \nabla_{\vS^{-1}} \myexpect_q \sqr{ f(\vm) + (\vm
        - \vparam)^\top \left. \nabla_{\vparam} f(\vparam) \right\vert_{\vtheta = \vm} + \half (\vm
        - \vparam)^\top \left. \nabla_{\vparam}^2 f(\vparam) \right\vert_{\vtheta = \vm} (\vm - \vparam) } \nonumber\\
   &= 2 \nabla_{\vS^{-1}} \trace \sqr{ \half \vS^{-1} \left. \nabla_{\vparam}^2 f(\vparam) \right\vert_{\vtheta = \vm}} \nonumber \\
    &= \left. \nabla_\vparam^2 f(\vparam) \right\vert_{\vparam=\vm}, 
      \label{eq:secondderiv}
\end{align}
where the first line is due to Price's theorem (\cref{app:bp}) and the second line is simply the second-order Taylor expansion.
We refer to this as the second-order delta method.

\updated{
\subsection{Deterministic Estimates as crude-Bayesians}
\label{eq:crudeBayesian}

In ML it is often argued that deterministic estimates/models can be seen as a special case of probabilistic ones where variables are set to their maximum-a-posterior (MAP) estimates. One justification stems from the fact that such estimates can be seen as a posterior which is a Dirac-delta distribution at the MAP estimate; see \citet{buntine2002variational, girolami2003equivalence, Gal2016Uncertainty} for examples. The argument is then to use the delta distribution as a solution to the Bayesian
objective (the evidence lower bound), justifying the MAP estimate as a Bayesian solution.

Formally, denoting the MAP estimate as $\vparam_*$, we can define a distribution $q_{\sigma}(\vparam) = \gauss(\vparam|\vparam_*, \sigma \vI)$ 
By taking $\sigma\to 0$, we can view $\elbofinal(q_{\sigma})$ as the Bayesian solution obtained by a non-Bayesian MAP estimate.
Unfortunately, this argument is flawed when estimating continuous parameters of deterministic models, as shown by \cite{welling2008deterministic}. This is because, as $\sigma\to 0$, the entropy $\mathcal{H}(q) = P\log \sigma \to -\infty$, making the objective useless. Therefore, it is not correct to use delta distributions to justify deterministic solutions as probabilistic and Bayesian ones.

Instead, we use the Delta method and obtain MAP estimates as a \emph{crude} Bayesian approximation of the expectation. In this method, we do not set the posterior as a Dirac delta, rather we use a posterior approximation with a pre-specified uncertainty. For example, we can use a Gaussian 
\[q_*(\vparam) = \gauss(\vparam| \vparam_*, \vH_*^{-1}) \] 
where $\vH_*$ is the Hessian of ${\barloss}(\vparam)$ or a positive-definite approximation of it when $\vH_*$ itself is not. This is the Laplace's method where the curvature around $\vparam_*$ is used as an uncertainty estimate.
The choice views the MAP estimate as a crude-Bayesian because, by using the Delta method, we can replace the expectation by loss at the mean $\vparam_*$,
\[
   \myexpect_{q}[\barloss(\vparam)] - \mathcal{H}(q) \,\, \approx \,\, \barloss(\vparam_*) + \half P \log |\vH_*| 
\]
which is exactly the objective optimized by $\vparam_*$ (the second term being a constant). A crude-Bayesian ignores the uncertainty around the estimate and simply uses the mean to approximate the expectation.

A similar view was presented in \citet{Opper:09} where Laplace's method was shown to be a crude-approximation of the variational method. However, in our argument the posterior covariance can be an arbitrary quantity, and even the posterior form does not need to be a Gaussian. The Delta method view is more general than the Laplace's view of \citet{Opper:09}.
}

\section{Newton's Method from the BLR}
\label{app:bp}

\updated{%
We start by expressing the natural gradients in terms of the gradient with respect to the mean and covariance. Let $\vparam$ be multivariate normal with mean $\vmu$ and covariance $\vC$, and $h(\vmu, \vC)$ be a differentiable function.  First, we write $\vm$ and $\vC$ in terms of $\vmeanparam^{(1)}$ and $\vmeanparam^{(2)}$,
\[
   \vm = \vmeanparam^{(1)}, \quad \quad \vC = \vmeanparam^{(2)} - \vmeanparam^{(1)}\rnd{\vmeanparam^{(1)}}^{\top},
\]
We then express the natural gradients in terms of gradients with respect to $\vm$ and $\vC$ by using the chain rule. For example, the derivative with respect to $\vmeanparam^{(2)}$ is as follows,
\[
   \deriv{h}{\meanparam^{(2)}_{ij}} = \sum_k \deriv{h}{m_k} \deriv{m_k}{\meanparam^{(2)}_{ij}} + \sum_{k,l} \deriv{h}{C_{kl}} \deriv{C_{kl}}{\meanparam^{(2)}_{ij}} = \deriv{h}{C_{kl}},
\]
giving us $\grad_{\vmeanparam^{(2)}} h = \grad_{\text{\vC}} h = \grad_{\text{\vS}^{-1}} h$. Similarly, the gradient with respect to $\vmeanparam^{(1)}$ is 
\begin{align*}
   \deriv{h}{\meanparam^{(1)}_i} &= \sum_k \deriv{h}{m_k} \deriv{m_k}{\meanparam^{(1)}_{i}} + \sum_{k,l} \deriv{h}{C_{kl}} \deriv{C_{kl}}{\meanparam^{(1)}_{i}} \\
   &= \deriv{h}{m_i} - \sum_k \deriv{h}{C_{ki}} \meanparam^{(1)}_{k} - \sum_l \deriv{h}{C_{il}} \meanparam^{(1)}_{l}\\
   &= \deriv{h}{m_i} - 2 \sum_k \deriv{h}{C_{ki}} m_{k},
\end{align*}
where the last line follows due to symmetry of $\vC$. Using this, we get the desired result,
\begin{align*}
   \grad_{\vmeanparam^{(1)}} h &= \grad_{\text{\vm}} h - 2[\grad_{\text{\vS}^{-1}} h] \vm, \\
   \grad_{\vmeanparam^{(2)}} h &= \grad_{\text{\vS}^{-1}} h
\end{align*}%
Next, we give }a summary of the Theorem's of Bonnet's and Price's
\citep{bonnet1964transformations, price1958useful}; see also
\citet{Opper:09, rezende2014stochastic}. Let $f(\vparam)$ be a twice
differentiable function where $\myexpect{|f(\vparam)|} < \infty$, then the
Bonnet and Price's results are
\begin{displaymath}
   \frac{\partial}{\partial\mu_i} \myexpect \sqr{ f(\vparam)} =
    \myexpect \sqr{ \frac{\partial}{\partial x_i} f(\vparam) }
    \quad\text{and}\quad
    \frac{\partial}{\partial C_{ij}} \myexpect \sqr{ f(\vparam) } =
    c_{ij}\myexpect \sqr{ \frac{\partial^{2}}{\partial x_i\partial x_j}f(\vparam) }
\end{displaymath}
where $c_{ij}=1/2$ if $i=j$ and $c_{ij}=1$ if $i\not=j$ (since $\vC$
is symmetric), respectively. Using these, we obtain
\cref{eq:ngrad_1_gauss} and \cref{eq:ngrad_2_gauss}.

The following derivation is also detailed in \cite{khan2018fast}. We
now use the definitions of natural and expectation parameters from
\cref{eq:full_gauss_params} in the BLR \cref{eq:bayes_learn_rule}.
The function $h(\vparam)$ is constant for the multivariate Gaussian, hence
\begin{align*}
  \vS_{t+1} \vm_{t+1}
  & \leftarrow (1-\rho_t) \vS_t\vm_t -
    \rho_t  \nabla_{\boldsymbol{\meanparam}^{(1)}}
    \myexpect_{q_t} [ \barloss(\vparam)]\\
  \vS_{t+1}
  & \leftarrow (1-\rho_t) \vS_t + 2
    \rho_t \nabla_{\boldsymbol{\meanparam}^{(2)}}
    \myexpect_{q_t} [ \barloss(\vparam)].
\end{align*}
The update for $\vS_{t+1}$ can be obtained using
\cref{eq:ngrad_2_gauss},
\begin{displaymath}
    \vS_{t+1} \leftarrow (1-\rho_t) \vS_t + \rho_t
    \myexpect_{q_t} \sqr{
        \nabla_{\text{\vparam}}^2 \barloss(\vparam)}.
\end{displaymath}
We can simplify the update for the mean using
\cref{eq:ngrad_1_gauss},
\begin{align*}
  \vS_{t+1} \vm_{t+1} &= (1-\rho_t) \vS_t\vm_t -
                        \rho_t  \crl{ \myexpect_{q_t}
                        [ \nabla_{\boldsymbol{\param}}
                        \barloss(\vparam)] -
                        \myexpect_{q_t} [
                        \nabla_{\boldsymbol{\param}}^2 \barloss(\vparam)] \vm_t }, \\
                      &= \vS_{t+1} \vm_t -
                        \rho_t  \myexpect_{q_t}
                        [ \nabla_{\boldsymbol{\param}} \barloss(\vparam)],
\end{align*}
and the update for $\vm_{t+1}$ follows from this.



\section{Natural-Gradient Updates for Mixture of Gaussians}
\label{app:mog}
%
%
%
The joint distribution $q(\vparam, z)$ is a minimal-conditional
exponential-family distribution \citep{linfast}, i.e., the distribution
then takes a minimal form conditioned over $z=k$ which ensures that the FIM is non-singular. Details of the update are in \citet[Theorem 3]{linfast}. The natural and expectation parameters are given in \citet[Table~1]{linfast}).

The gradients with respect to $\vmu_k$ can be written in terms of
$\vm_k$ and $\vS_k^{-1}$ as
\begin{align*}
  \nabla_{\boldsymbol{\meanparam}_k^{(1)}}
  \myexpect_{q}
  [\barloss(\vparam)]
  &= \sqr{
    \nabla_{\boldsymbol{\meanparam}_k^{(1)}}
    \vm_k^\top }
    \nabla_{\boldsymbol{\mathbf{m}}_k}
    \myexpect_{q}
    [\barloss(\vparam)] = \frac{1}{\pi_k}
    \nabla_{\boldsymbol{\mathbf{m}}_k} \myexpect_{q}
    [\barloss(\vparam)]
\end{align*}
Hence we can utilize from the derivation of Newton's method,
\cref{eq:ngrad_1_gauss} and \cref{eq:ngrad_2_gauss}, and do
similarly as in \cref{app:bp}, to achieve a Newton-like update
for a mixture of Gaussians
\begin{align}
  \vm_{k,t+1} &\leftarrow \vm_{k,t} - \frac{\rho}{\pi_k}
                \vS_{k,t+1}^{-1} \nabla_{\mathbf{m}_k}
                \myexpect_{q_t} \sqr{
                \barloss(\vparam) + \log q(\vparam)},
                \label{eq:exp-von10}\\
  \vS_{k,t+1} &\leftarrow \vS_{k,t} +
                \frac{2\rho}{\pi_{k}} \nabla_{\mathbf{S}_{k,t}^{-1}}
                \myexpect_{q_t} \sqr{
                \barloss(\vparam) + \log q(\vparam)}.
                \label{eq:exp-von20}
\end{align}
The gradients can be expressed in terms of the gradient and Hessian of
the loss, similar to Bonnet's and Price's Theorem in
\cref{app:bp}, see \cite[Thm~6 and 7]{lin2019stein}. This gives
\begin{align*}
  \nabla_{\mathbf{m}_k} \myexpect_{q}[f(\vparam)]
  &= \pi_k \myexpect_{\mathcal{N}(\text{\vparam}|\mathbf{m}_{k},
    \mathbf{S}_{k}^{-1})} [ \nabla_{\boldsymbol{\param}} f(\vparam)] , \\
  \nabla_{\mathbf{S}_k^{-1}}
  \myexpect_{q}[f(\vparam)]
  &=  \frac{\pi_k}{2}
    \myexpect_{\mathcal{N}(\text{\vparam}|\mathbf{m}_{k},
    \mathbf{S}_{k}^{-1})} [ \nabla_{\boldsymbol{\param}}^2 f(\vparam)] ,
\end{align*}
and hence the updates in \cref{eq:exp-von1,eq:exp-von2}.

The gradient of
$\log q(\vparam) = \log \sum_{k=1}^K \pi_k \gauss(\vparam|\vm_k,
\vS_k^{-1})$ were not spelled out explicit in
\cref{eq:exp-von10,eq:exp-von20}, but are as follows,
\begin{align}
  \nabla_{\text{\vparam}} \log q(\vparam)
  &= \sum_{j=1}^K  r_j(\vparam)  \vS_j (\vm_j - \vparam), \label{eq:mixexp_1st} \\
   \nabla^2_{\text{\vparam}} \log q(\vparam) 
  &= \sum_{j=1}^K r_j(\vparam) \sqr{\vA_{jj}(\vparam)
    -\vS_j - \sum_{i=1}^K r_i(\vparam) \vA_{ij}(\vparam) }. \label{eq:mixexp_2nd} 
\end{align}
Here, $r_k(\vparam)$ is the responsibility of the $k$'th mixture
component at $\vparam$, $q(z=k|\vparam)$
\begin{align}
  q(z=k|\vparam) = \frac{\pi_{k} \gauss(\vparam|\vm_{k}, \vS_{k}^{-1})}{
  \sum_{j=1}^K \pi_{j} \gauss(\vparam|\vm_{j}, \vS_{j}^{-1})} .
\end{align}
and
$\vA_{ij}(\vparam) = \vS_i (\vm_i-\vparam) (\vm_j-\vparam)^\top
\vS_j$.

\updated{With two natural assumptions, we can get simplifications that help us understand how different components can take responsibility for different local minima.} First, we assume 
 $\vparam_k = \vm_k$, then 
 $\vA_{ik}(\vparam_k) = 0$ for all $i$, leading to 
 \begin{align}
   \nabla_{\text{\vparam}} \log q(\vparam_k)
   &= \sum_{j\ne k}  r_j(\vparam_k)  \vS_j (\vparam_j - \vparam_k), \label{eq:mixexp_1st_1} \\
   \nabla^2_{\text{\vparam}} \log q(\vparam_k)
   &= - r_k(\vparam_k)\vS_k + \sum_{j\ne k } r_j(\vparam_k)
     \sqr{  \vA_{ii}(\vparam_k) - \vS_j  -
     \sum_{i=1, i\ne k}^K r_i(\vparam_k) \vA_{ij}(\vparam_k) } ,\label{eq:mixexp_2nd_2} 
 \end{align}
 Second, we assume the mixture components to be far apart from each other, then
 the responsibility is approximately zero for all $j\ne k$, leading to
 \begin{align}
   \nabla_{\text{\vparam}} \log q(\vparam_k)
   &\approx 0, \quad\quad 
     \nabla^2_{\text{\vparam}\text{\vparam}} \log q(\vparam_k) \approx - \vS_k.
     \label{eq:gaussmix_assumption}
 \end{align}



\section{Deep Learning Algorithms with Momentum}
\label{sec:dl_momentum}

Momentum is a useful technique to improve the performance of SGD
\citep{sutskever2013importance}. Modern adaptive-learning algorithms,
such as RMSprop and Adam, often employ variants of the classical
momentum to increase the learning rate
\citep{sutskever2013importance}.

The classical momentum method is based on the Polyak's heavy-ball
method \cite{polyak1964some}, where the current
$\vparam_{t+1} -\vparam_t$ update is pushed along the direction of the
previous update $\vparam_t - \vparam_{t-1}$:
\begin{align}
   \vparam_{t+1} \leftarrow \vparam_t - \alpha \widehat{
  \nabla}_{\mathbf{\param}} \barloss(\vparam_t) + \gamma
  (\vparam_t - \vparam_{t-1})  
  \label{eq:sgd_momentum}
\end{align}
with a fixed momentum coefficient $\gamma>0$. Why this technique can
be beneficial when using noisy gradient is more transparent when we
rewrite \cref{eq:sgd_momentum} as
\begin{equation}
    \vparam_{t+1} \leftarrow \vparam_t -
    \alpha \sum_{k=0}^t \gamma^{k}\widehat{
        \nabla}_{\mathbf{\param}} \barloss(\vparam_{t-k}) +
    \text{initial conditions}.
\end{equation}
Hence, the effective gradient is an average of previous gradients with
decaying weights. The hope is that this will smooth out and reduce the variance of the
stochastic gradient estimates, although exact mechanisms behind the success of momentum-based methods are still an open research question.

Adaptive-learning rate algorithms, such as RMSprop and Adam, employ a
variant of the classical momentum method where exponential-smoothing
is applied to the gradient
$\vu_{t+1} \leftarrow \gamma \vu_t + (1-\gamma) \widehat{
    \nabla}_{\mathbf{\param}} \barloss(\vparam_t)$ which is used in
the update while keeping the scaling unchanged
\citep{graves2013generating, kingma2014adam}. This gives the update
\begin{align}
  \vparam_{t+1} \leftarrow \vparam_t - \alpha
  \frac{1}{\sqrt{\vs_{t+1}} + c \vone} \circ \vu_{t+1}.
  \label{eq:adam_momentum}  
\end{align}
We can express this similar to \cref{eq:sgd_momentum}, as
\citep{wilson2017marginal}
\begin{align}
  \vparam_{t+1} &\leftarrow \vparam_t - \alpha(1-\gamma)
                  \frac{1}{\sqrt{\vs_{t+1}} +
                  c \vone} \circ \sqr{\widehat{
                  \nabla}_{\mathbf{\param}}
                  \barloss(\vparam_t) } + \gamma\frac{\sqrt{\vs_{t}} +
                  c \vone}{\sqrt{\vs_{t+1}} +
                  c \vone} \circ (\vparam_t -\vparam_{t-1}),
   \label{eq:adam_moentum_alternate}
\end{align}
showing a dynamic scaling of the momentum depending (essentially) on
the ratio of $\vs_{t+1}$ and $\vs_t$. Although this adaptive scaling
is not counter-intuitive, it is a result of the somewhat arbitrary
choices made. Can we justify this from the Bayesian principles and
derive the corresponding BLR? The problem we face is provoked by
the use of stochastic gradient approximations and the principles
simply state that we should compute exact gradients instead. This
might appear as a deficiency in our main argument but we can resolve
this issue in other ways.

It is our view that the natural way to include a momentum term in the
BLR, is to do it within the mirror descent framework discussed in
\cref{sec:bayes_learn}. This will result in an update where the
momentum term obey the geometry of the posterior approximation. We
propose to augment \cref{eq:md_mu} with a momentum term
\begin{align}
  \vmeanparam_{t+1} \leftarrow \arg\min_{\boldsymbol{\meanparam} \in
  \mathcal{M}}\,\, \myang{ \nabla_{\boldsymbol{\meanparam}}
  \mathcal{L}(q_t) , \vmeanparam}
  + \frac{1+\gamma_t}{\rho_t} \mathbb{D}_{A^*}(\vmeanparam \|
  \vmeanparam_t)
  - \frac{\gamma_t}{\rho_t}
  \mathbb{D}_{A^*}(\vmeanparam \| \vmeanparam_{t-1})
   \label{eq:md_momentum}
\end{align}
The last term penalizes the proximity to $\vmu_{t-1}$ where the
proximity is measured according to the KL divergence, following the
suggestion by \citet{khan2018fast} we may interpret this as the
`natural momentum' term. This gives a (revised) BLR with
momentum,
\begin{align}
   \vnatparam_{t+1} \leftarrow \vnatparam_t -
        \rho_t \natgrad_\vnatparam \sqr{ 
        \myexpect_{q_t}
        \rnd{\barloss(\vparam)} - \entropy(q_t) }
                     + \gamma_t (\vnatparam_t - \vnatparam_{t-1}).
   \label{eq:bayes_learn_rule_momentum}
\end{align}
This update will recover \cref{eq:sgd_momentum} and an alternative to
\cref{eq:adam_moentum_alternate}. Choosing a Gaussian candidate
distribution $\gauss(\vparam|\vm,\vI)$ (the result is invariant to a
fixed covariance matrix), we recover \cref{eq:sgd_momentum} after a
derivation similar as in \cref{sec:grad_desc}. Choosing the candidate
distribution $\gauss(\vparam|\vm,\diag(\vs)^{-1})$, follow again
previous derivations (Newton's method in \cref{sec:blr_newton}) with
mini-batch gradients, and result in the following update
\begin{equation}
    \begin{split}
        \vparam_{t+1} &\leftarrow \vparam_t -
        \rho_t \frac{1}{\vs_{t+1}} \circ
        \sqr{ \widehat{\nabla}_{\mathbf{\param}} 
            \barloss(\vparam_t) } +
        \frac{1}{\vs_{t+1}} \circ
        ( \vs_t \vparam_t - \vs_{t-1} \vparam_{t-1}), \\  
        \vs_{t+1} &\leftarrow (1-\rho_t) \vs_t +
        \rho_t\,\,
        \diag\sqr{ \widehat{\nabla}_{\mathbf{\param}\mathbf{\param}}^2 
            \barloss(\vparam_t) }  + \gamma_t (\vs_t - \vs_{t-1}).
    \end{split}
    \label{eq:don_momentum_1}
\end{equation}
Assuming $\vs_t \approx \vs_{t-1}$ the last terms in the above two
equations simplifies, and then by replacing $\vs_t$ by $\sqrt{s}_t$
and adding a constant $c$, we can recover
\cref{eq:adam_moentum_alternate}. This momentum-version of
\cref{eq:don}, not only justifies the use of scaling vectors for the
momentum, but also the use of exponential-smoothing for the scaling
vectors itself.



\section{Dropout from the BLR}
\label{app:dropout}

Deriving updates for $\vm_{jl}$ and $\vS_{jl}$ with dropout is similar
to the Newton variant derived for mixture of Gaussian discussed in
\cref{app:mog}. With dropout there are only two components
where one of the component has fixed parameters, hence one set of
natural and expectation parameters per weight vector $\vparam_{jl}$,
\begin{equation*}
    \begin{array}{ll}
      \begin{array}{ll} 
        \vnatparam_{jl}^{(1)} &= \vS_{jl}\vm_{jl}, \\
        \vnatparam^{(2)}_{jl} &=-\half \vS_{jl}, 
      \end{array}
     \begin{array}{ll} 
       \quad\vmeanparam_{jl}^{(1)}
       & = \myexpect_{q}
         \sqr{ 1_{[\lat_{jl}=1]} \vparam_{jl}} = \pi_1 \vm_{jl} , \\
       \quad\vmeanparam_{jl}^{(2)}
       & = \myexpect_{q}
         \sqr{ 1_{[\lat_{jl}=1]} \vparam_{jl}\vparam_{jl}^T }
         = \pi_1 \rnd{ \vS_{jl}^{-1} + \vm_{jl}\vm_{jl}^T }, 
     \end{array}
    \end{array}
\end{equation*}
We can reuse the update \cref{eq:exp-von10,eq:exp-von20} from
\cref{app:mog}
\begin{align}
  \vm_{jl,t+1} &\leftarrow \vm_{jl,t} -
                 \frac{\rho}{\pi_1} \vS_{jl,t+1}^{-1}
                 \nabla_{\mathbf{m}_{jl}}
                 \crl{ \myexpect_{q_t}
                 \sqr{ \barloss(\vparam)} +
                 \myexpect_{q_t}
                 \sqr{ \log q(\vparam_{jl})} } ,\label{eq:von_dropout_1_0}\\
  \vS_{jl,t+1} &\leftarrow \vS_{jl,t+1} +
                 \frac{2\rho}{\pi_1}
                 \nabla_{\mathbf{S}_{jl}^{-1}}
                 \crl{ \myexpect_{q_t}
                 \sqr{ \barloss(\vparam)} +
                 \myexpect_{q_t}
                 \sqr{ \log q(\vparam_{jl})} }.\label{eq:von_dropout_2_0}
\end{align}
We have moved the expectation inside the gradient since the entropy
term only depends on $\vparam_{jl}$. The gradient of the loss
$\barloss(\vparam)$ may depend on the whole $\vparam$, however.

By using appropriate methods to compute gradients, we can recover the
dropout method. Specifically, for the loss term, we use
the reparameterization trick \citep{kingma2013auto} and approximate the
gradients at the samples from $q(\vparam_{jl})$, 
\begin{displaymath}
    \vparam_{jl} = z_{jl} (\vm_{jl} + \vS_{jl}^{-1/2} \vepsilon_{1,jl}) +
    (1-z_{jl})s_0^{-1/2} \vepsilon_{2,jl} 
\end{displaymath}
where $z_{jl}$ is a sample from Bernoulli distribution with
probability $\pi_1$, and $\vepsilon_{1,jl}$ and $\vepsilon_{2,jl}$ are
two independent samples from standard normal distribution. Since
weights are deterministic in dropout, we need to ignore
$\vepsilon_{1,jl}$ and $\vepsilon_{2,jl}$ to recover dropout from the
BLR. The gradients are then evaluated at the dropout mean
$\tvm_{jl} = z_{jl} \vm_{jl}$ (the delta method) which leads to the following gradient
approximations:
\begin{equation}
    \begin{split}
        \nabla_{\mathbf{m}_{jl}} \myexpect_{q_t} [
        \barloss(\vparam) ] &\approx \left. \nabla_{\vparam_{jl}}
     \barloss(\tvparam) \right\vert_{\tvparam = \tvm_t}, \\
        2 \nabla_{\mathbf{S}_{jl}^{-1}}
        \myexpect_{q_t} \sqr{ \barloss(\vparam)} & =
        \nabla_{\mathbf{m}_{jl}\mathbf{m}_{jl}^\top}^2
        \myexpect_{q_t} \sqr{ \barloss(\vparam)}
        \approx \left. \nabla_{\tvparam_{jl}\tvparam_{jl}^\top}^2
        \barloss(\tvparam) \right\vert_{\tvparam = \tvm_t}.
    \end{split}
   \label{eq:grad_loss}
\end{equation}
We will use these gradients for the loss term in
\cref{eq:von_dropout_1_0,eq:von_dropout_2_0}.

Gradients of $\log q(\vparam_{jl})$ only depends on the variables
involved in $q(\vparam_{jl})$, and we can use the delta method similar
to \cref{eq:mog_delta} in \cref{sec:opt}. We make the
assumption that the two mixture components are ``not close'' which is
reasonable with dropout. With this assumption then the gradient
respect to $\vm_{jl}$ is approximately zero
\begin{align}
  \nabla_{\mathbf{m}_{jl}} \myexpect_{q}
  \sqr{ \log q(\vparam_{jl})}
  &= \nabla_{\mathbf{m}_{jl}} \int d\vparam_{jl}
    \log q(\vparam_{jl}) \sqr{ \pi_1 \gauss(\vparam_{jl}| \vm_{jl},
    \vS_{jl}^{-1}) + (1-\pi_1) \gauss(\vparam_{jl}| \vzero, s_0^{-1}
    \vI_{n_l}) } \nonumber\\ 
  &= \pi_1 \nabla_{\mathbf{m}_{jl}} \int d\vparam_{jl}
    \log q(\vparam_{jl}) \gauss(\vparam_{jl}| \vm_{jl},
    \vS_{jl}^{-1}) \nonumber\\ 
  &= \pi_1 \myexpect_{\mathcal{\mathcal{N}}(\text{\vparam}_{jl}|\mathbf{m}_{jl,t},
    \mathbf{S}_{jl,t}^{-1})}
    \sqr{ \nabla_{\mathbf{\param}_{jl}}
    \log q(\vparam_{jl})} \text{ (Using Bonnet's Theorem)}\nonumber\\
  &\approx \pi_1 \nabla_{\mathbf{m}_{jl}}
    \log q(\vm_{jl,t}) \text{ (Using the Delta approximation) } \nonumber\\
  &\approx 0. \label{eq:grad_logq}
\end{align}
For the gradient with respect to $\vS^{-1}$ we get
\begin{align}
  2 \nabla_{\mathbf{S}_{jl}^{-1}} \myexpect_{q}
  \sqr{ \log q(\vparam_{jl})} 
  &= \nabla_{\mathbf{m}_{jl}\mathbf{m}_{jl}^\top}^2
    \myexpect_{q} \sqr{ \log q(\vparam_{jl})} \nonumber\\
  &= \nabla_{\mathbf{m}_{jl}}
    \sqr{ \pi_1  \myexpect_{\mathcal{\mathcal{N}}(\text{\vparam}_{jl}|\mathbf{m}_{jl,t},
    \mathbf{S}_{jl,t}^{-1})}
    \sqr{ \nabla_{\mathbf{\param}_{jl}}
    \log q(\vparam_{jl})} } \text{ (using first derivative wrt $\mathbf{m}_{jl}$) }\nonumber\\
  &= \pi_1 \myexpect_{\mathcal{\mathcal{N}}(\mathbf{\param_{jl}}|\mathbf{m}_{jl,t},
    \mathbf{S}_{jl,t}^{-1})}
    \sqr{
    \nabla_{\mathbf{\param}_{jl}\mathbf{\param}_{jl}^\top}^2
    \log q(\vparam_{jl})} \text{ (Using Bonnet's theorem) } \nonumber\\
  &\approx \pi_1
    \nabla_{\mathbf{m}_{jl}\mathbf{m}_{jl}^\top}^2
    \log q(\vm_{jl,t}) \text{ (Using Delta approximation) }\nonumber\\
  &\approx -\pi_1 \vS_{jl,t},
    \label{eq:hess_logq}
\end{align}
Using the gradient approximations
\cref{eq:grad_loss,eq:grad_logq,eq:hess_logq} in
\cref{eq:von_dropout_1_0,eq:von_dropout_2_0}, we get the
\begin{align*}
  \vm_{jl,t+1} &\leftarrow \vm_{jl,t} -
                 \frac{\rho}{\pi_1} \vS_{jl,t+1}^{-1}
                 \nabla_{\mathbf{m}_{jl}}
                 \crl{ \nabla_{\mathbf{m}_{jl}} \barloss(\tvm_t) +  0} ,\\
  \vS_{jl,t+1} &\leftarrow \vS_{jl,t+1} +
                 \frac{\rho}{\pi_1}
                 \nabla_{\mathbf{S}_{jl}^{-1}}
                 \crl{ \nabla_{\mathbf{m}_{jl}\mathbf{m}_{jl}^\top}^2
                 \barloss(\tvm_t)
                 - \pi_1 \vS_{jl,t} }.
\end{align*}
which reduces to the updates \cref{eq:von_dropout_1,eq:von_dropout_2}.



\section{Table of exponential-family distributions}

\begin{table}[!ht]
   \begin{center}
      \begin{tabular}{p{1in} c c c p{0.7in}}
         \toprule
         Distribution & $\vnatparam$ & $\vT(\vparam)$ & $\vmu$ & Reference\\
         \bottomrule
          Gaussian $\gauss(\vparam|\vm, \vI)$ with fixed covariance & $\vm$ & $\vparam$ & $\vm$ & \cref{sec:grad_desc} \\
         \bottomrule
          Gaussian $\gauss(\vparam|\vm, \vS^{-1})$ with fixed covariance & $\vS\vm$ & $\vparam$ & $\vm$ & \cref{sec:sgd}\\
         \bottomrule
          Gaussian $\gauss(\vparam|\vm, \vS^{-1})$ with diagonal $\vS = \diag(\vs)$ & $\rnd{ \begin{array}{c} \vm \\ -\half\vs \end{array} }$ & $\rnd{ \begin{array}{c} \vparam \\ \vparam \circ\vparam \end{array} }$ & $\rnd{ \begin{array}{c} \vm\\ \vm\circ\vm + 1/\vs \end{array} }$ & \cref{sec:ada_dl,sec:uncertainty_dl} \\
         \bottomrule
          Gaussian $\gauss(\vparam|\vm, \vS^{-1})$ & $\rnd{ \begin{array}{c} \vm \\ -\half\vS \end{array} }$ & $\rnd{ \begin{array}{c} \vparam \\ \vparam\vparam^\top \end{array} }$ & $\rnd{ \begin{array}{c} \vm\\ \vm\vm^\top + \vS^{-1} \end{array} }$ & \cref{sec:blr_newton,sec:ridge} \\
         \bottomrule
          Finite mixture of Gaussian $P(z=k, \vparam) = \pi_k\gauss(\vparam|\vm_k, \vS_k^{-1})$ with fixed $\pi_k$& $\rnd{ \begin{array}{c} \vm_k \\ -\half\vS_k \end{array} }$ & $\rnd{ \begin{array}{c} \mathbf{1}_{[z=k]}\vparam \\ \mathbf{1}_{[z=k]}\vparam\vparam^\top \end{array} }$ & $\rnd{ \begin{array}{c} \pi_k\vm_k\\ \pi_k\rnd{\vm_k\vm_k^\top + \vS_k^{-1} } \end{array} }$ & \cref{sec:dropout,sec:multimodal} \\
         \bottomrule
          Bernoulli with $P(\param = 1) = p$ & $\half \log \frac{p}{1-p}$ & $2 \mathbf{1}_{[\theta=k]} - 1$ & $2p-1$ & \cref{sec:binn} \\
         \bottomrule
      \end{tabular}
   \end{center}
   \caption{\updated{A summary of exponential-family used in the paper showing natural parameters $\vnatparam$, sufficient statistics $\vT(\vparam)$, and expectation parameters $\vmeanparam$, along with the relevant sections.}}
   \label{tab:EFsummary}
\end{table}

\newpage
\bibliography{refs,hrue}

\end{document}